\documentclass[12pt]{article}
\usepackage{amsmath}
\usepackage{times}
\pdfoutput=1
\usepackage{graphicx}
\usepackage{color}
\usepackage{multirow}
\usepackage{longtable}
\usepackage{psfrag}
\usepackage[authoryear]{natbib}
\usepackage{multirow}
\usepackage{rotating}
\usepackage{bbm}
\usepackage{latexsym}
\usepackage{epstopdf}
\usepackage{caption}
\usepackage{subfig} %should come after caption package if given, can't be used with caption2, defines subfloat
\newcommand{\captionfonts}{\normalsize}

\makeatletter

\makeatletter

\makeatletter  
\long\def\@makecaption#1#2{%
  \vskip\abovecaptionskip
  \sbox\@tempboxa{{\captionfonts #1: #2}}%
  \ifdim \wd\@tempboxa >\hsize
    {\captionfonts #1: #2\par}
  \else
    \hbox to\hsize{\hfil\box\@tempboxa\hfil}%
  \fi
  \vskip\belowcaptionskip}
\makeatother   
\begin{document}
\hspace{13.9cm}1

\ \vspace{20mm}\\

{\LARGE An Online Structural Plasticity Rule for Generating Better Reservoirs}

\ \\
{\bf Subhrajit Roy}\\
\emph{subhrajit.roy@ntu.edu.sg}\\
\emph{School of Electrical and Electronic Engineering, Nanyang Technological University, Singapore.}\\
{\bf Arindam Basu}\\
\emph{arindam.basu@ntu.edu.sg}; Corresponding author\\
\emph{School of Electrical and Electronic Engineering, Nanyang Technological University, Singapore.}\\

%\ \\[-2mm]

\thispagestyle{empty}
\markboth{}{NC instructions}
\ \vspace{-0mm}\\
%
%Abstract
\begin{center} {\bf Abstract} \end{center}
In this article, a novel neuro-inspired low-resolution online unsupervised learning rule is proposed to train the reservoir or liquid of Liquid State Machine. The liquid is a sparsely interconnected huge recurrent network of spiking neurons. The proposed learning rule is inspired from structural plasticity and trains the liquid through formation and elimination of synaptic connections. Hence, the learning involves rewiring of the reservoir connections similar to structural plasticity observed in biological neural networks. The network connections can be stored as a connection matrix and updated in memory by using Address Event Representation (AER) protocols which are generally employed in neuromorphic systems. On investigating the `pairwise separation property' we find that trained liquids provide 1.36 $\pm$ 0.18 times more inter-class separation while retaining similar intra-class separation as compared to random liquids. Moreover, analysis of the `linear separation property' reveals that trained liquids are 2.05 $\pm$ 0.27 times better than random liquids. Furthermore, we show that our liquids are able to retain the `generalization' ability and `generality' of random liquids. A memory analysis shows that trained liquids have 83.67 $\pm$ 5.79 $ms$ longer fading memory than random liquids which have shown 92.8 $\pm$ 5.03 $ms$ fading memory for a particular type of spike train inputs. We also throw some light on the dynamics of the evolution of recurrent connections within the liquid. Moreover, compared to `Separation Driven Synaptic Modification' - a recently proposed algorithm for iteratively refining reservoirs, our learning rule provides 9.30\%, 15.21\% and 12.52\% more liquid separations and 2.8\%, 9.1\% and 7.9\% better classification accuracies for four, eight and twelve class pattern recognition tasks respectively.       

\section{Introduction}
In neural networks, `plasticity' of synapses refers to their connection rearrangements and changes in strengths over time. Over the past decade, a plethora of learning rules have been proposed which are capable of training networks of spiking neurons through various forms of synaptic plasticity \citep{gardner2015,Kuhlmann2014,Sporea2013,Chronotron,ReSuMe_neco,Gutig2006,Brader2007,Arthur2006,Gerstner2002,Moore2002,Mel2001}. A large majority of these works have explored weight plasticity where a neural network is trained by modifying (strengthening or weakening) of the synaptic strengths. We identify another unexplored form of plasticity mechanism termed as structural plasticity that trains a neural network through formation and elimination of synapses. Since structural plasticity involves changing the network connections over time, it does not need the provision to keep high-resolution weights. Hence, it is inherently a low resolution learning rule. Such a type of low-resolution learning rule is motivated by the following biological observations:

\begin{enumerate}
	\item Biological experiments have shown that the strength of synaptic transmission at cortical synapses can experience considerable fluctuations ``up" and ``down" representing facilitation and depression respectively, or both, when excited with short synaptic stimulation and these dynamics are distinctive of a particular type of synapse \citep{Thomson1993,Markham1996,Varela1997,Hempel2000}. This kind of short-time dynamics is in contrary to the traditional connectionist models assuming high-resolution synaptic weight values and conveys that synapses may only have a few states.
	\item Moreover, another experimental research on long-term potentiation (LTP) in the hippocampus region of the brain revealed that excitatory synapses may exist in only a small number of long-term stable states, where the continuous grading of synaptic efficacies observed in common measures of LTP may exist only in the average over a huge population of low-resolution synapses with randomly staggered thresholds for learning \citep{Petersen1998}.  
\end{enumerate}

Since the learning happens through formation and elimination of synapses, structural plasticity has been used to train networks of neurons with active dendrites and binary synapses \citep{Mel2001,Hussain2014_nc,roy_biocas1,roy_tbcas,Roy_MOT_tnnls_2015}. These works demonstrated that networks constructed of neurons with nonlinear dendrites and binary synapses and trained through structural plasticity rules can obtain superior performance for supervised and unsupervised pattern recognition tasks.
 
However, till now structural plasticity has been employed as a learning rule only in the context of neurons with non-linear dendrites and binary synapses. We identify that it is a generic rule and can be tailored to train any neural network thereby allowing it to evolve in low-resolutions. We intend to venture into this domain and as a first step we have chosen to modify the sparse recurrent connections of a liquid or reservoir of Liquid State Machine (LSM) \citep{Maass2002} constructed of standard Leaky Integrate and Fire (LIF) neurons through structural plasticity and study its effects. In our algorithm, structural plasticity happening in longer timescales is guided by a fitness function updated by a STDP inspired rule in shorter timescales. In contrast to the recently proposed algorithms that aim to enhance the LSM performance by evolving its liquid \citep{Xue2013,Hourdakis2013,Notley2012,Obst2012}, our structural plasticity-based training mechanism provides the following advantages:
\begin{enumerate}
	\item For hardware implementations, the `choice' of connectivity can be easily implemented exploiting Address Event Representation (AER) protocols commonly used in current neuromorphic systems where the connection matrix is
	stored in memory. Unlike traditional algorithms modifying and storing real-valued synaptic weights, online learning in real time scenarios is achieved by the proposed algorithm through only modification of the connection table stored in memory.    
	\item Due to the presence of positive feedback connections in liquid, training it through `weight plasticity' might lead to an average increase in synaptic weight and eventually take it to the unstable region. We conjecture that since our connection based learning rule keeps the average synaptic weight of the liquid constant throughout learning, it reduces the chance of leading the liquid into instability.  
\end{enumerate}  

The rest of the paper is organized as follows. In the following section, we shall discuss the LSM framework and throw some light on the previous works on LSM and structural plasticity. In Section \ref{unsup_str_plas}, we will propose the  structural plasticity based unsupervised learning rule we have developed to train a liquid or reservoir composed of spiking neurons. In Section \ref{exp_res}, we shall share and discuss the results obtained from the experiments performed to evaluate the performance of our liquid. We will conclude the paper by discussing the implications of our work and future directions in the last section. We have included an Appendix section where we list the specification of the liquid architecture and values of parameters used in this article.

\section{Background and Theory}
In this section, we shall briefly describe the LSM framework and look into few works from the last few years that aspired to improve the liquid or reservoir of LSM. Next we shall briefly review few supervised and unsupervised structural plasticity learning rules.

\subsection{Theory of Liquid State Machine}
\begin{figure}[!t]
	\begin{center}
	\includegraphics[width=0.95\textwidth]{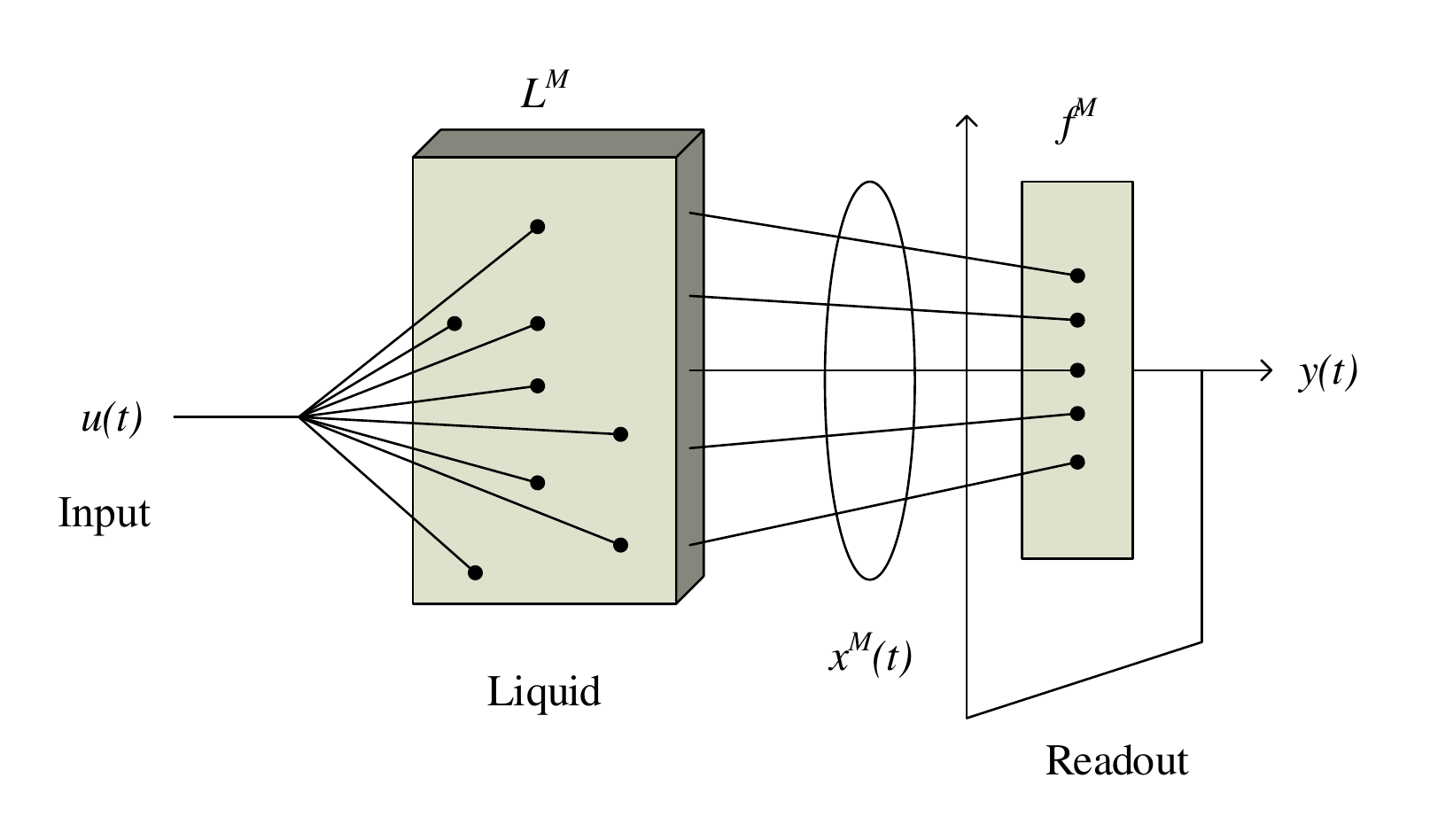}
	\caption{In this figure the LSM framework is shown. LSM consists of three stages with the first stage being the input layer. The input stage is followed by a pool of recurrent spiking neurons whose synaptic connections are generated randomly and are usually not trained. The next stage is a simple linear classifier that is selected and trained in a task-specific manner.}
	\label{fig:LSM}
	\end{center}
\end{figure}

LSM \citep{Maass2002} is a reservoir computing method developed from the viewpoint of computational neuroscience by Maass et al. It supports real-time computations by employing a high dimensional heterogeneous dynamical system which is continuously perturbed by time-varying inputs. The basic structure of LSM is shown in Fig. \ref{fig:LSM}. It comprises three parts: an input layer, a reservoir or liquid, and a memoryless readout circuit.  The liquid is a recurrent interconnection of a large number of biologically realistic spiking neurons and synapses. The readout is implemented by a pool of neurons which do not possess any lateral interconnections. The spiking neurons of the liquid are connected to the neurons of the readout. The liquid does not create any output, instead it transforms the lower dimensional input stream to a higher dimensional internal state. These internal states act as an input to the memoryless readout circuit which is responsible for producing the final output of LSM.

Following \citep{Maass2002}, if $u(t)$ is the input to the reservoir, then the liquid neuron circuit can be represented mathematically as a liquid filter $L^M$ which maps the input function $u(t)$ to the internal states $x^M(t)$ as:
\begin{equation}
\label{eq1}
x^M(t)=(L^Mu)(t)
\end{equation}

The next part of LSM i.e. the readout circuit takes these liquid states as input and transforms them at every time instant $t$ into the output $y(t)$ given by:
\begin{equation}
\label{eq2}
y(t)=f^M(x^M(t))
\end{equation}

The liquid circuit is general and does not depend on the problem at hand whereas the readout is selected and trained in a task-specific manner. Moreover, multiple readouts can be used in parallel for extracting different features from the internal states produced by the liquid. For more details on the theory and applications of LSM, we invite the reader to refer to \citep{Maass2002}.

\subsection{Previous research on improvement of liquid}\label{research_liquid}
In \citep{Xue2013}, a novel Spike Timing Dependent Plasticity (STDP) based learning rule for generating a self-organized liquid was proposed. The authors showed that LSM with STDP learning rule provides better performance than LSM with randomly generated liquid. \citet{Ju2013} have studied in detail the effect of the distribution of synaptic weights and synaptic connectivity in the liquid on LSM performance. In addition, the authors have proposed a genetic algorithm based rule for the evolution of the liquid from a minimum structure to an optimized kernel having an optimal number of synapses and high classification accuracy. The authors of \citep{Hourdakis2013} have used the Fisher’s Discriminant Ratio as a measure of the Separation Property of the liquid. Subsequently, they have used this measure in an evolutionary framework to generate liquids with suitable parameters, such that the performance of readouts get optimized. \citet{Sillin2013} implemented a reservoir by using Atomic Switch Networks (ASN) that is capable of utilizing nonlinear dynamics without needing to control or train the connections in the reservoir. They showed a method to optimize the physical device parameters to maximize efficiency for a given task. \citet{Schliebs2012} have proposed an algorithm to dynamically vary firing threshold of the liquid neurons in presence of neural spike activity such that LSM is able to achieve both a high sensitivity of the liquid to weak inputs as well as an enhanced resistance to over-stimulation for strong stimuli. \citet{Wojcik2012} simulated the LSM by forming the liquid with Hodgkin-Huxley neurons instead of LIF neurons as used in \citep{Maass2002}. They have done a detailed analysis of the influence of cell electrical parameters on the Separation Property of this Hodgkin-Huxley liquid. In \citep{Notley2012}, Notley et al. have proposed a learning rule which updates the synaptic weights in the liquid by using a tri-phasic STDP rule. \citet{Frid2012} have proposed a modified version of LSM which can successfully approximate real-valued continuous functions. Instead of providing spike trains as input to the liquid, they have directly provided the liquid with continuous inputs. Moreover, they have also used neurons with firing history dependent sliding threshold to form the liquid. In \citep{Hazan2012}, the authors have shown that LSM in its normal form is less robust to noise in data but when certain biologically plausible topological constraints are imposed then the robustness can be increased. The authors of \citep{Schliebs_ijcnn_2011} have presented a liquid formed with stochastic spiking neurons and termed their framework as pLSM. They have shown that due to the probabilistic nature of the proposed liquid, in some cases pLSM is able to provide better performance than traditional LSM. In \citep{RheAume2011}, the authors have proposed novel techniques for generating liquid states to improve the classification accuracy. First, they have presented a state generation technique which combines the membrane potential and firing rates of liquid neurons. Second, they have suggested representing the liquid states in frequency domain for short-time signals of membrane potentials. Third, they have shown that combination of different liquid states lead to better performance of the readout. \citet{Kello2010} have presented a self-tuning algorithm that is able to provide a stable liquid firing activity in the presence of a varied range of inputs. A self-tuning algorithm is used with the liquid neurons that adjusts the post-synaptic weights in such a way that the spiking dynamics remain between sub-critical and super-critical. In \citep{Norton2010}, the authors have proposed a learning rule termed as `Separation Driven Synaptic Modification (SDSM)' for training the liquid, which is able to construct a suitable liquid in fewer generations than random search.   

\subsection{Previous research involving Structural Plasticity}

\citet{Mel2001} showed that supervised classifiers employing neurons with active dendrites i.e. dendrites having lumped non-linearities and binary synapses can be trained through structural plasticity to recognize high-dimensional Boolean inputs. Inspired by \citep{Mel2001}, \citet{Hussain2014_nc} proposed a supervised classifier composed of neurons with non-linear dendrites and binary synapses suitable for hardware implementations. In \citep{roy_biocas1,roy_tbcas} this classifier was upgraded to accommodate spike train inputs and a spike-based structural plasticity rule was proposed that was amenable for neuromorphic implementations. In \citep{Roy_MOT_tnnls_2015} another supervised spike-based structural plasticity rule inspired from Tempotron \citep{Gutig2006} was proposed for training a threshold-adapting neuron with active dendrites and binary synapses. Moreover, an unsupervised spike-based strutural plasticty learning rule was proposed in \citep{Roy_unsup_2015} for training a Winner-Take-All architecture composed of neurons with active dendrites and binary synapses. 

Till now, the works related to structural plasticity that we showcased in this subsection have been confined to networks employing neurons with nonlinear dendrites and binary synapses. In recent years, structural plasticity has been used to train generic neural networks as well. For example, in \citep{George2015a}, a STDP learning rule interlaced with structural plasticity was proposed to train feed-forward neural networks. In \citep{George2015b}, this work was modified and applied to a highly recurrent spiking neural network. In these two works, the synaptic weights are modified by STDP and when a critical period is reached, the synapses are pruned through structural plasticity. Unlike these works, that implement an interplay between STDP and structural plasticity, in the proposed learning rule structural plasticity or connection modifications happen on longer timescales (at the end of patterns) which is guided by the fitness function or correlation coefficient updated by a STDP inspired rule in shorter timescales (at each pre- and post-synaptic spike). In the following section, we will propose our connection based unsupervised learning rule.

\section{Unsupervised structural plasticity in liquid} \label{unsup_str_plas}
\begin{figure} [!t]
	\centering 
	\includegraphics[width=0.95\textwidth, height = 7.5cm]{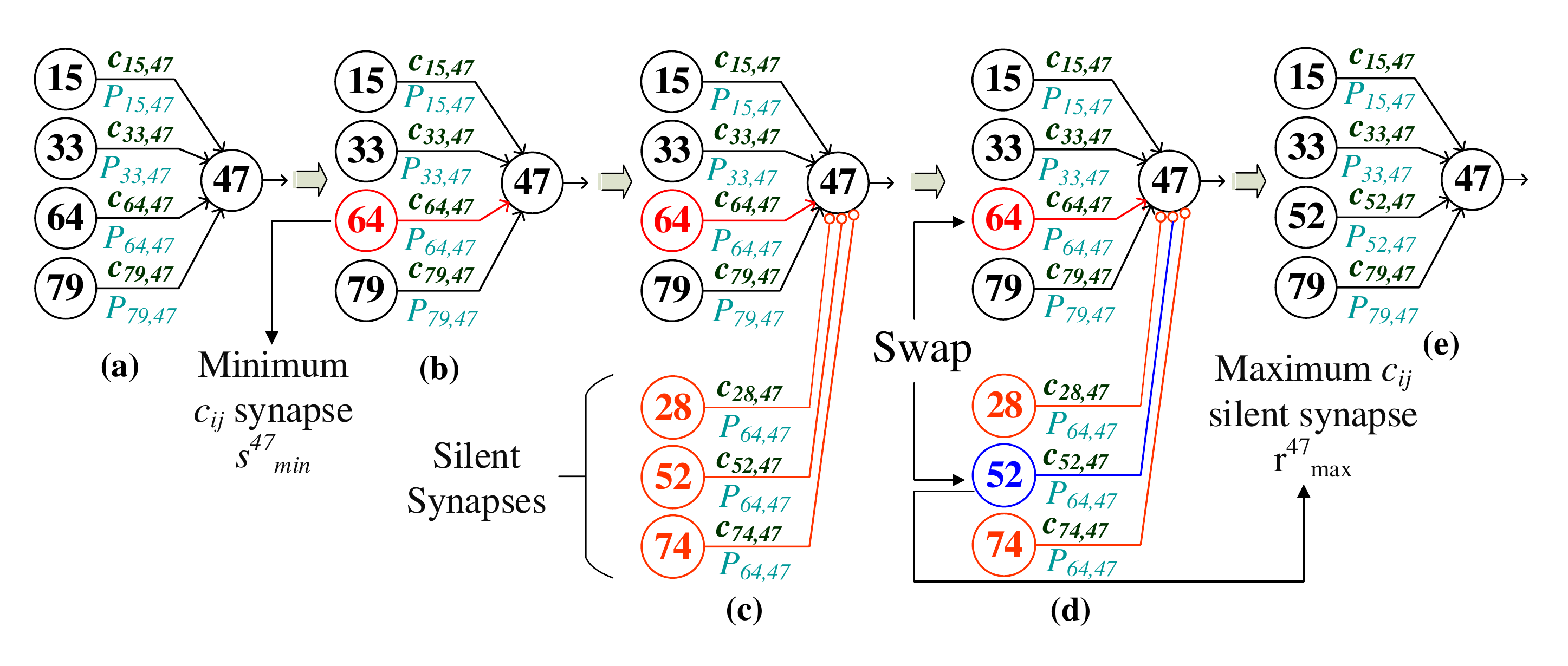}
	\caption{(Caption in the following page.)}
	\label{fig:swapping}
\end{figure}

The liquid is a three-dimensional recurrent architecture of spiking neurons that are sparsely connected with synapses of random weights. We used Leaky Integrate and Fire (LIF) neurons and biologically realistic dynamic synapses to construct our liquids. The specifications of the architecture and the parameter values are listed in the Appendix and unless otherwise mentioned, we use these values in our experiments. In the proposed algorithm, learning happens through formation and elimination of synapses instead of the traditional way of updating real-valued weights associated with them. Hence, to guide the unsupervised learning, we define a correlation coefficient based fitness value $c_{ij}(t)$ for the synapse (if present) connecting the $j^{th}$ excitatory neuron to the $i^{th}$ excitatory neuron, as a substitute for its weight. Since the liquid neurons are sparsely connected, $c_{ij}(t)$'s are defined only for the neuron pairs that have synaptic connections. Note that the proposed learning rule only modifies the connections between excitatory neurons. The operation of the network and the learning process comprises the following steps whenever a pattern is presented:  

\begin{itemize}
	\item  $c_{ij}(t)$ is initialized as $c_{ij}(t=0)=0 \; \forall$ excitatory neuron pairs $(i,j)$ having a synaptic connection from the $j^{th}$ to the $i^{th}$ neuron. Here, $L_e$ is the number of excitatory liquid neurons and $i\in{0,1,...L_e} \; \& \; j\in{0,1,...L_e}$. Note that $L$ is the total number of neurons in the liquid which comprises of $L_e$ excitatory and $L_i$ inhibitory neurons.
	\item  The value of $c_{ij}(t)$ is depressed at pre-synaptic and potentiated at post-synaptic spikes according to the following rule:
	\begin{enumerate}
		\item Depression: If a post-synaptic spike of the $j^{th}$ excitatory neuron appears after some time delay $\delta_{ij}$ as a pre-synaptic spike to the $i^{th}$ excitatory neuron at time $t^{pre}$, then the value of $c_{ij}(t)$ is updated by a quantity $\Delta c_{ij}(t^{pre})$ given by:
		\begin{equation}
		\Delta c_{ij}(t^{pre})= -\;\bar{f}_{i}(t^{pre})
		\end{equation}
		where $f_i(t)$ and $\bar{f}_{i}(t)= K(t) \ast f_i(t)$ are the output and the post-synaptic trace of the $i^{th}$ excitatory spiking neuron. In this work we have chosen an exponential form of $K(t)$ given by $K(t)= I_0 (e^{-t/\tau_s}-e^{-t/\tau_f})$.
		\item Potentiation: If the $i^{th}$ excitatory neuron fires a post-synaptic spike at time $t^{post}$ then $c_{ij}(t)$ for each synapse connected to it is updated by $\Delta c_{ij}(t^{post})$ given by:
		\begin{equation}
		\Delta c_{ij}(t^{post})=\bar{e}_{j}(t^{post})
		\end{equation}
		where $\delta_{ij}$ and $\bar{e}_{j}(t) = K(t) \ast f_{j}(t-\delta_{ij})$ are its delay and the pre-synaptic trace of the spiking input it receives.
	\end{enumerate}
	A pictorial explanation of this update rule of $c_{ij}(t)$ is shown in Fig. \ref{fig:pre_post_pre}.
	\begin{figure}[!t]
		\centerline{
			\includegraphics[width=0.9\textwidth]{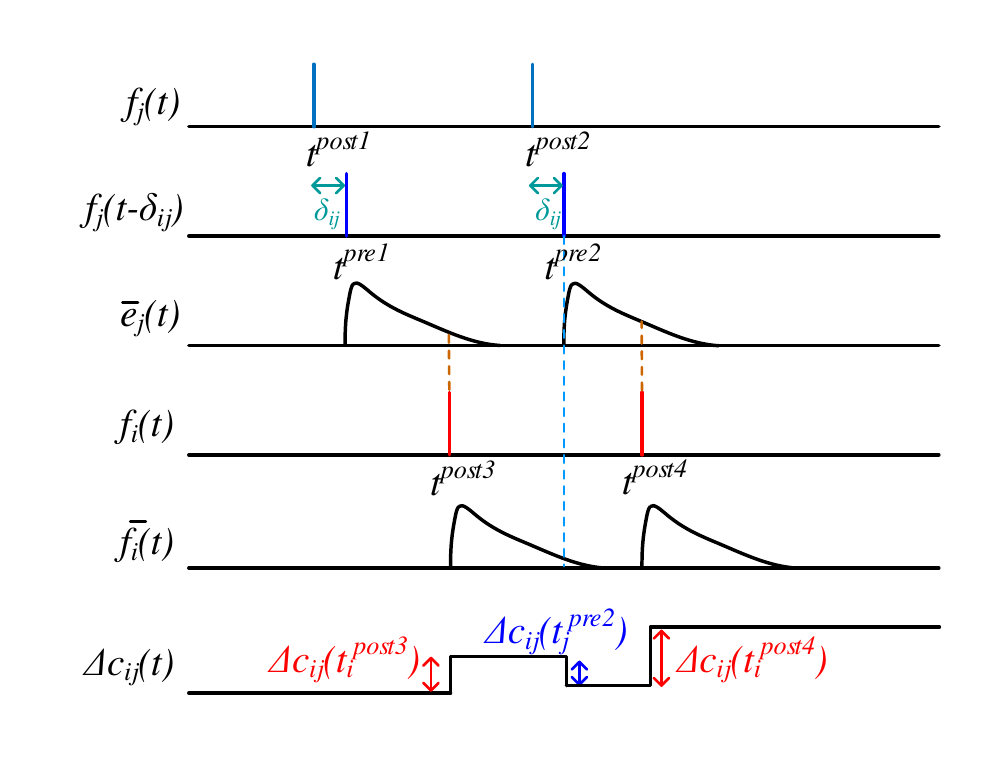}}
		\caption{An example of the update rule of fitness value ($c_{ij}(t)$) associated with the synapse connecting the $j^{th}$ excitatory neuron to the $i^{th}$ excitatory neuron is shown. When a post-synaptic spike is emitted by excitatory neuron $i$ at $t^{post3}$, the value of $c_{ij}(t)$ increases by $\bar{e}_{j}(t^{post3})$. On the other hand, when excitatory neuron $j$ emits a post-synaptic spike at $t^{post2}$, it reaches neuron $i$ at $t^{pre2} = t^{post2} + \delta_{ij}$ due to the presence of synaptic delay. The arrival of a pre-synaptic spike at $t^{pre2}$ reduces $c_{ij}(t)$ by $\bar{f}^{i}(t^{pre2})$ as shown in the figure.}
		\label{fig:pre_post_pre}
	\end{figure}
	\item After the network has been integrated over the current pattern, the synaptic connections of the excitatory neurons which have produced at least one spike are modified. 
	\item If we consider that $Q$ out of $L_e$ excitatory neurons have produced a post-synaptic spike for the current pattern, then the synaptic connections of the $q^{th}$ neuron $\forall \; q=1,2...,Q$ is updated by tagging the synapse ($s_{min}^{q}$) having the lowest value of correlation coefficient out of all the synapses connected to it, for possible replacement.    
	\item To aid the unsupervised learning process, randomly chosen sets $R^{q}$ containing $n_R$ of the $L_e$ excitatory neurons are forced to make connections to the $q^{th}$ neuron through silent synapses having same physiological parameters as $s_{min}^{q} \; \forall \; q=1,2...,Q$. We term these synapses as ``silent" since they do not contribute to the computation of the neuron's membrane voltage $V^n(t)$ - so they do not alter the liquid output when the same pattern is re-applied. The value of $c_{qj}(t)$ is calculated for synapses in $R^{q}$ and $s_{min}^{q}$ is replaced with the synapse having maximum $c_{qj}(t)$ ($r_{max}^{q}$) in $R^{q}$ $\forall q=1,2...,Q$ i.e. the pre-synaptic neuron connected to $s_{min}^{(q)}$ is swapped with the pre-synaptic neuron connected to $r_{max}^{q}$.
	\item All the $c_{ij}$ values are reset to zero and the above-mentioned steps are repeated for every pre- and post-synaptic spike. The proposed learning rule is explained by demonstrating the connection modification of a single neuron in Fig. \ref{fig:swapping}.
\end{itemize}  

\addtocounter{figure}{-1}
\begin{figure} [t!]
	\caption{In this figure, the proposed learning rule is explained by showing an example of synaptic connection modification. After the presentation of a pattern, a set of excitatory neurons $Q$ produces output spikes. Although connection modification takes place for all the neurons in set $Q$, the $47^{th}$ excitatory neuron is randomly chosen for this example and all the stages of its connection modification are shown in (a)-(e). The $15^{th}$, $33^{rd}$, $64^{th}$ and $79^{th}$ excitatory neurons  are connected to the $47^{th}$ excitatory neuron as shown in (a). Above each connection or synapse we mention the $c_{i,j}$ associated with that synapse. Moreover, below each synapse we mention the set of physiological parameters $P_{i,j} = \{w_{ij}, \delta_{ij}, \tau_{ij}, U_{ij}, D_{ij}, F_{ij}\}$ associated with it which includes the weight ($w_{ij}$), delay ($\delta_{ij}$), synaptic time constant ($\tau_{ij}$) and the $U_{ij}$, $D_{ij}$ and $F_{ij}$ parameters. Note that the $U_{ij}$, $D_{ij}$ and $F_{ij}$ parameters are only applicable for dynamic synapses. The $c_{ij}$ values of all the connections are checked at the end of the pattern and the one having the minimum value ($s_{min}^{47}$) is identified. In this example the connection from the $64^{th}$ excitatory neuron is identified as the minimum $c_{ij}$ connection and tagged for replacement (b). In the next step (c) a randomly chosen set of excitatory neurons are forced to make connections to the $47^{th}$ neuron through silent synapses. Note that the physiological parameters for these silent synapses are set to $P_{64,47}$ i.e. same as the synapse connecting the $64^{th}$ excitatory neuron to the $47^{th}$ excitatory neuron. The next step involves identifying the silent synapse having the maximum value of $c_{ij}$ i.e. $r_{max}^{47}$ which is the connection from the $52^{nd}$ excitatory neuron. Subsequently, our learning rule swaps these connections to form the updated morphology as shown in ((d)-(e)). \\}
\end{figure}

\section{Experiments and results} \label{exp_res}
In this section, we will describe the experiments performed to evaluate the performance of our algorithm and discuss the obtained results. We will employ various metrics to explore different properties of our liquid and compare it with the traditionally used randomly generated liquid. Moreover, we will compare its performance with another algorithm that performs iterative refining of liquids.  

\subsection{Separation capability} \label{sep_cap}
\begin{figure} [!t]
	\centering 
	\subfloat[]{\includegraphics[width=0.5\textwidth]{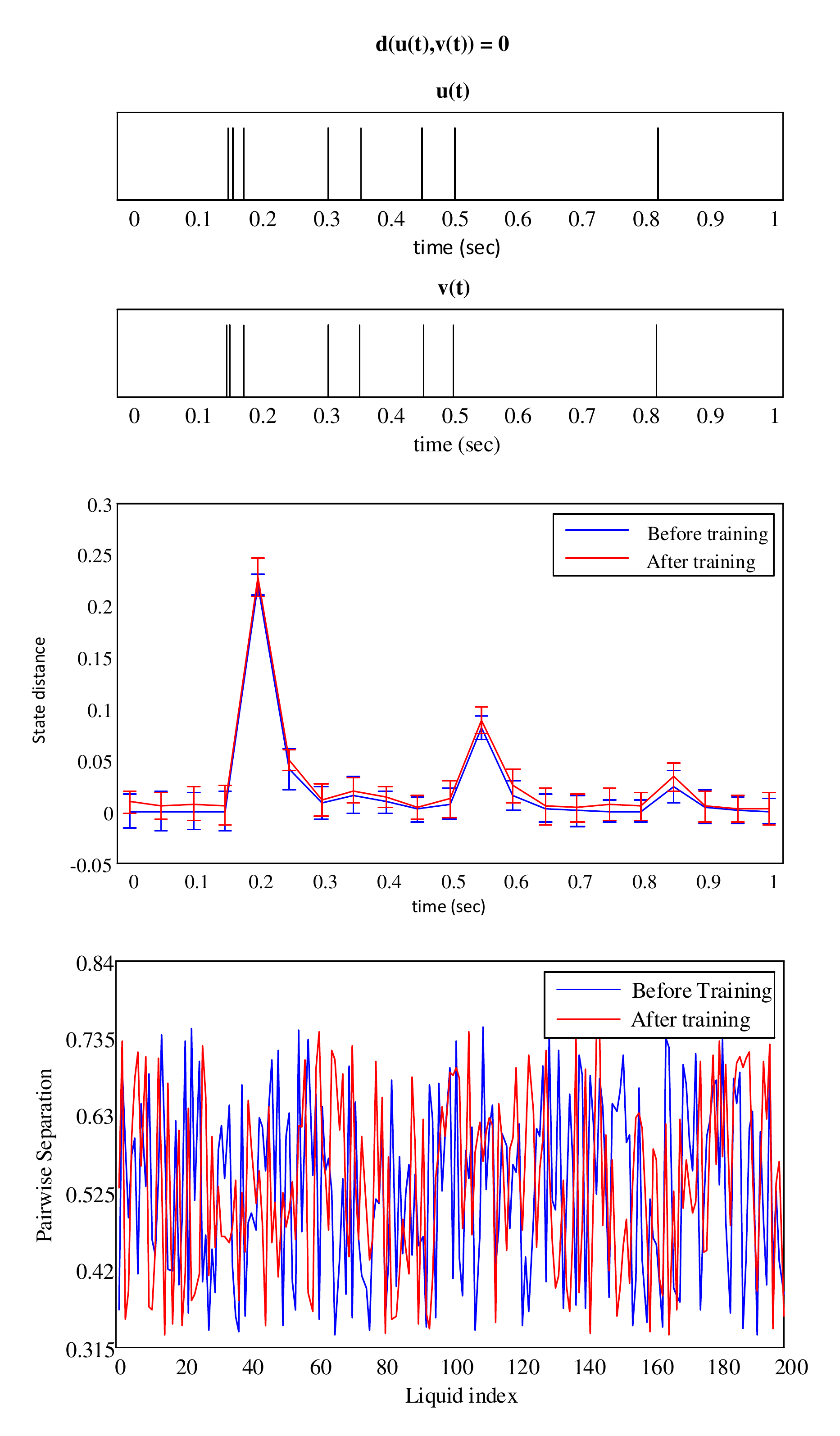}}
	\subfloat[]{\includegraphics[width=0.5\textwidth]{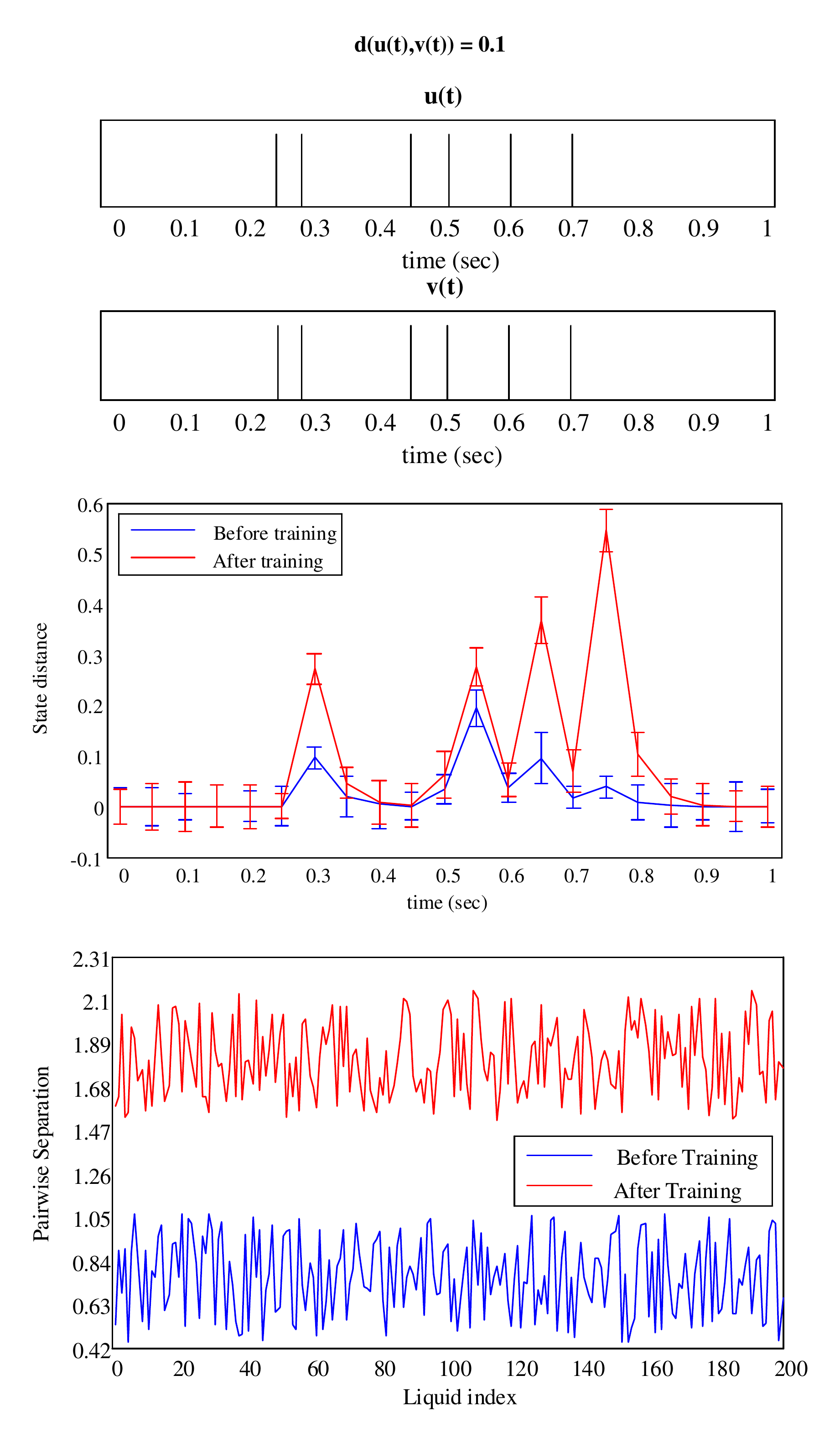}}
	\caption{(Caption in the following page.)}
	\label{pairwise1}
\end{figure}

In this sub-section, we throw some light on the computational power of our liquid trained by structural plasticity by evaluating its separation capability on spike train inputs. As a first test, we choose the \emph{pairwise separation property} considered in \cite{Maass2002} as a measure of its kernel quality. First, we take a spike train pair $u(t)$ and $v(t)$ of duration $T_p = 1$ sec and give it as input to numerous randomly generated liquids having different initial conditions in each trial. In this experiment, while the inputs remain same, the liquids are different for different trials. The resulting internal states of the liquid $x_u^M(t)$ and $x_v^M(t)$ are noted for $u(t)$ and $v(t)$ respectively. We calculate the average Euclidean norm $||x_u^M(t)-x_v^M(t)||$ between these two internal states and plot them in Fig. \ref{pairwise1} and \ref{pairwise2} against time for various fixed values of distance $d(u(t),v(t))$ between the two injected spike trains $u(t)$ and $v(t)$. To compute the distance between $u(t)$ and $v(t)$ we use the same method proposed in \citep{Maass2002}. Moreover, we show the Pairwise Separation obtained at each trial at the output of the liquid which is given by the following formula:
\begin{equation}
\textrm{Pairwise Separation} = \sum\limits_{n \;\;s.t. \;\; 0 < t_n < T_p} ||x_u^M(t_n)-x_v^M(t_n)||  
\end{equation}	 
where the liquid output is sampled at times $t_n$. 
\addtocounter{figure}{-1}
\begin{figure} [t!]
	\caption{Same input, different liquid: In this figure, we compare the \emph{pairwise separation property} of randomly generated liquids and the same liquids when trained through the proposed unsupervised structural plasticity based learning rule. While the blue lines indicate the separation capability of randomly generated liquids, the red lines are obtained by evolving the same liquids by the proposed algorithm. For this simulation, in each trial the connections are updated for 15 iterations. In each iteration one connection update happens for each spike train of the pair and hence the liquid encounters 30 connection modifications during training. In this figure, (a) shows that the noise level of both the random and the learnt liquid is similar. Hence, during training our learning rule does not induce extra noise. Moreover, (b) depicts that liquid trained by structural plasticity is able to obtain more separation than random liquids for $d(u(t),v(t)=$ 0.1 for all the trials. \\}
\end{figure}

\begin{figure} [!t]
	\centering 
	\subfloat[]{\includegraphics[width=0.5\textwidth]{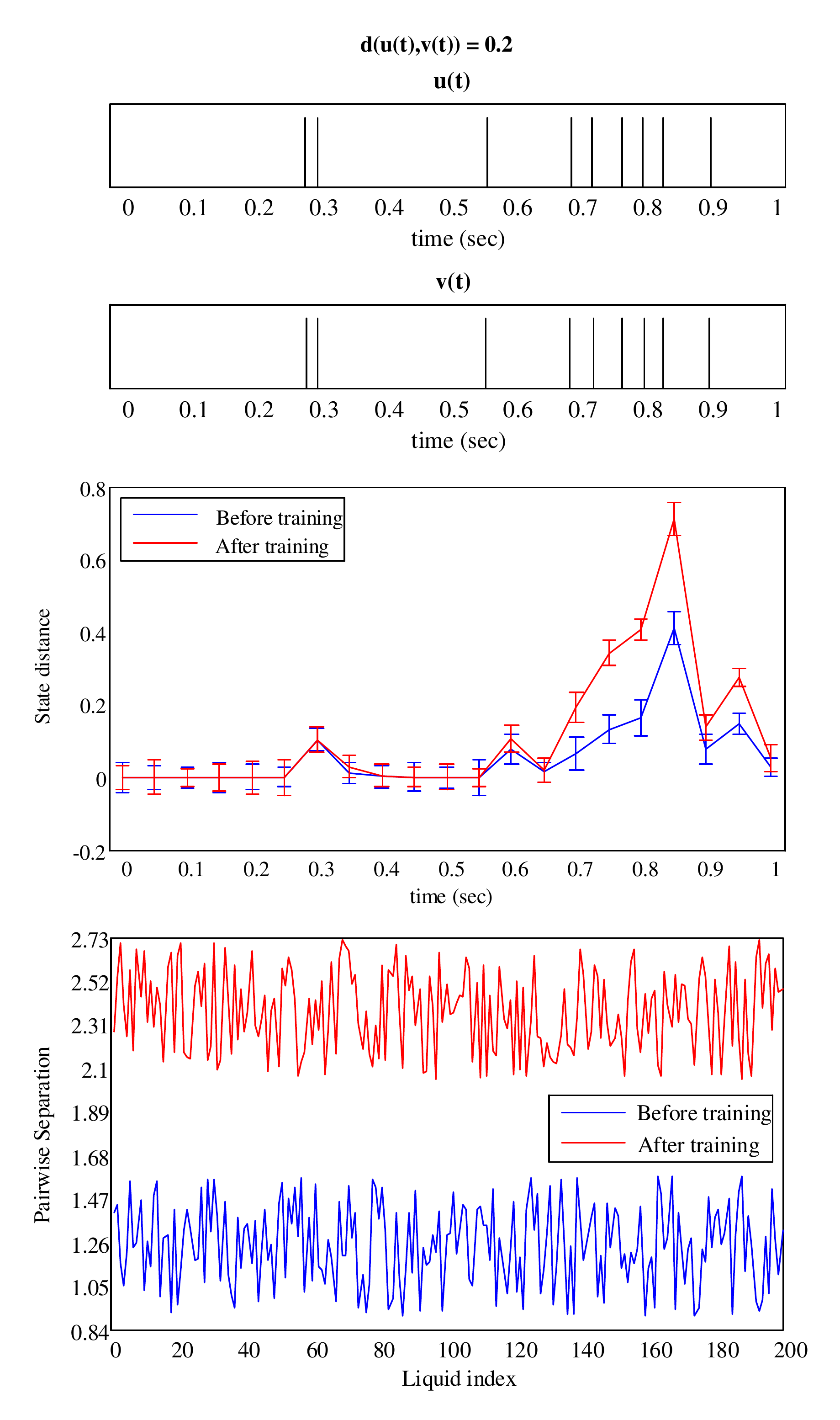}}
	\subfloat[]{\includegraphics[width=0.5\textwidth]{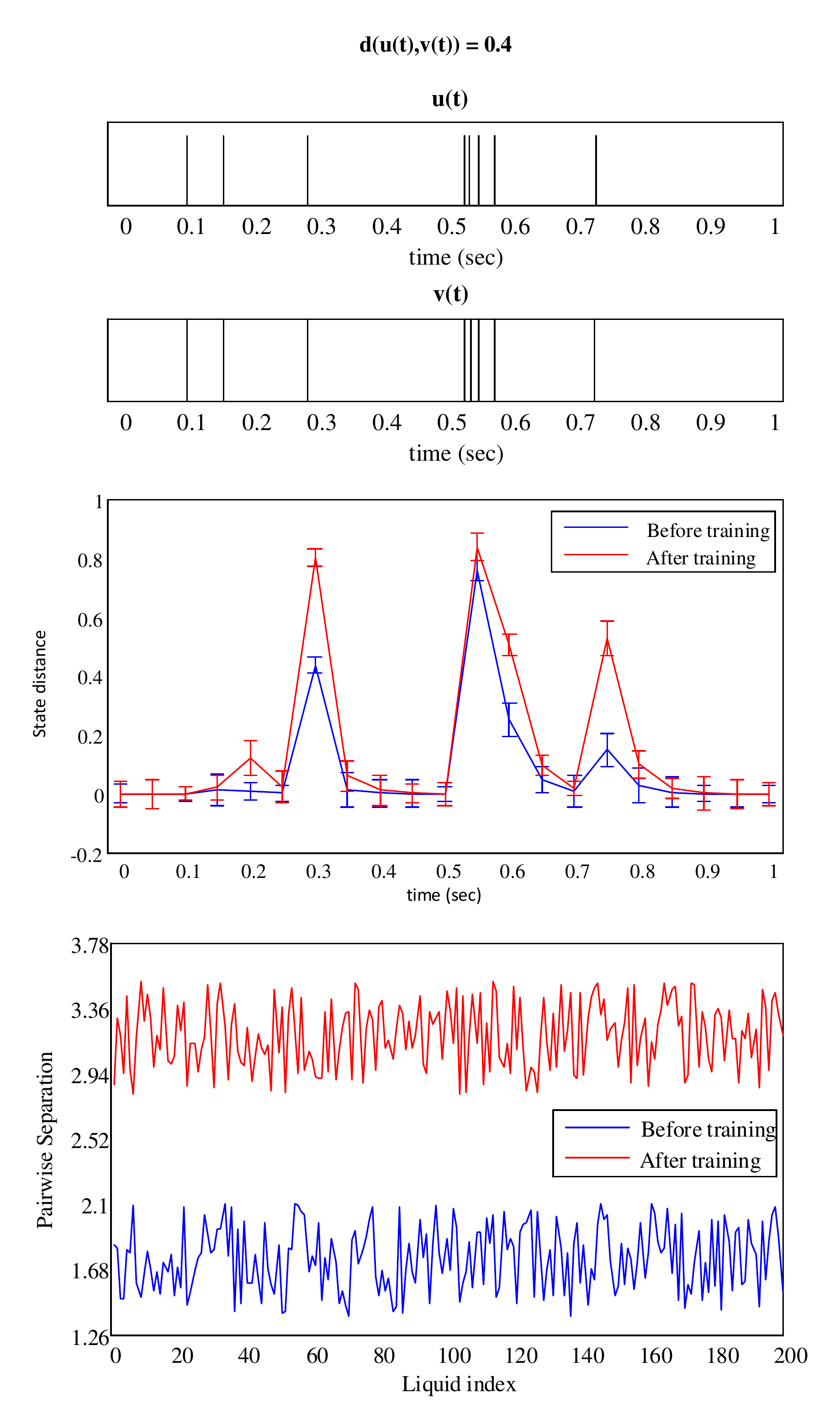}}
	\caption{Same input, different liquid: In this figure, the experiment performed to generate Fig. \ref{pairwise1} has been repeated for input spike train pairs of distance $d(u_(t),v_(t)=$ 0.2 (a) and 0.4 (b). Similar to Fig. \ref{pairwise1}(b) the trained liquids demonstrate more separation for all trials.}
	\label{pairwise2}
\end{figure}

In Fig. \ref{pairwise1} and  Fig. \ref{pairwise2}, the blue curves correspond to the traditional randomly generated liquid with no evolution as proposed in \cite{Maass2002} and the red curves correspond to the scenario when the same randomly generated liquid (for each trial) is trained through structural plasticity. Hence, the blue curves indicate the initial condition for generation of the red curves. In Fig. \ref{pairwise1}(a) i.e. the curves for $d(u(t),v(t)) = 0$ are generated by applying the same spike train to two randomly chosen initial conditions of the liquid. Hence, these curves show the noise level of the liquid. Fig. \ref{pairwise1}(b), Fig. \ref{pairwise2}(a) and Fig. \ref{pairwise2}(b) correspond to $d(u(t),v(t)) = 0.1$, $d(u(t),v(t)) = 0.2$ and $d(u(t),v(t)) = 0.4$ respectively. The parameters of the liquids are kept as shown in Table \ref{params_table} in the Appendix . Fig. \ref{pairwise1}(a) depicts that the separation of the liquid states are similar for both the methods. This demonstrates that the noise level is same for both the liquids. On the other hand, Fig. \ref{pairwise1} (b) and  Fig. \ref{pairwise2}(a)-(b) show that the proposed algorithm is able to produce liquid states with more separation than the traditional liquid with no evolution. In other words, morphologically evolved liquids are better in capturing and amplifying the separation between input spike trains as compared to randomly generated liquids. 

\begin{figure}[!t]
	\begin{center}
		\includegraphics[width=0.7\textwidth]{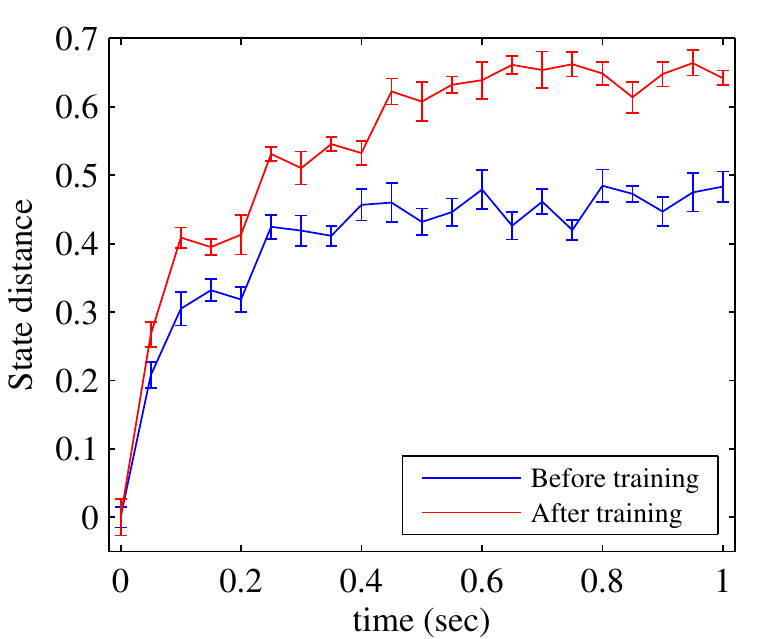}
		\caption{Same liquid, different input: In this figure, the same liquid is excited by 500 different input spike trains and the separation between the internal state distances are recorded. The resulting state distance, averaged over 500 trials, is plotted for both the randomly generated liquid and the same liquid when trained through structural plasticity. It is clear that the trained liquid always produces more separation than the random liquid at its output.}
		\label{same_liq_diff_in}
	\end{center}
\end{figure}

In the next experiment, we use the same liquid for all trials but vary the input spike pairs. 500 different spike train pairs $u_i(t)$ and $v_i(t)$ (here $i$ varies from 1 to 500) of distance $d(u_i(t),v_i(t)) = 0.2$ are generated and given as input to the same liquid separately. The internal state distance separation for all the trials are averaged and plotted in Fig. \ref{same_liq_diff_in} against time for both the random and trained liquids. The input spike trains typically tend to produce peaks and troughs in the state distance plot. They appear at different regions for different input spike trains. However, these peaks and troughs average out in Fig. \ref{same_liq_diff_in} since we take the mean of state distances obtained from 500 trials with different inputs. This figure clearly depicts that a liquid trained through structural plasticity consistently shows higher separation than a random liquid for different inputs.

\begin{figure} [!t]
	\centering 
	\subfloat[]{\includegraphics[width=0.48\textwidth]{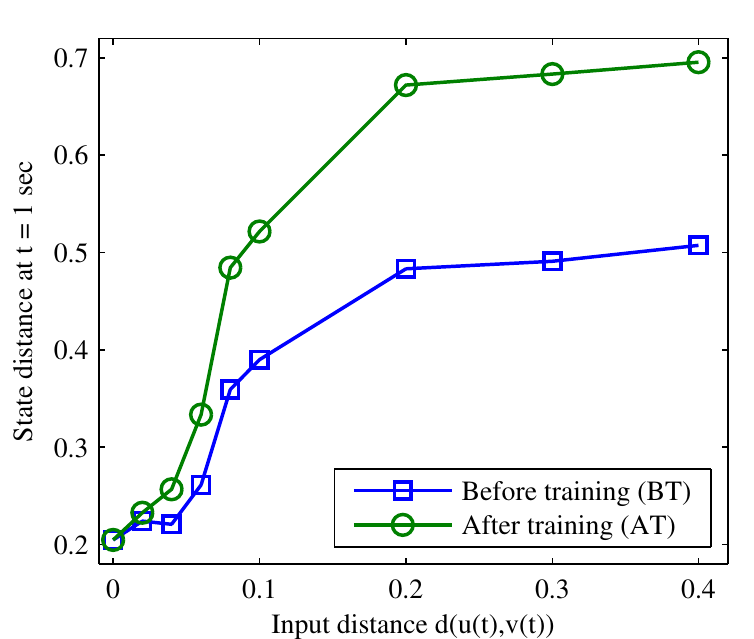}}
	\subfloat[]{\includegraphics[width=0.48\textwidth]{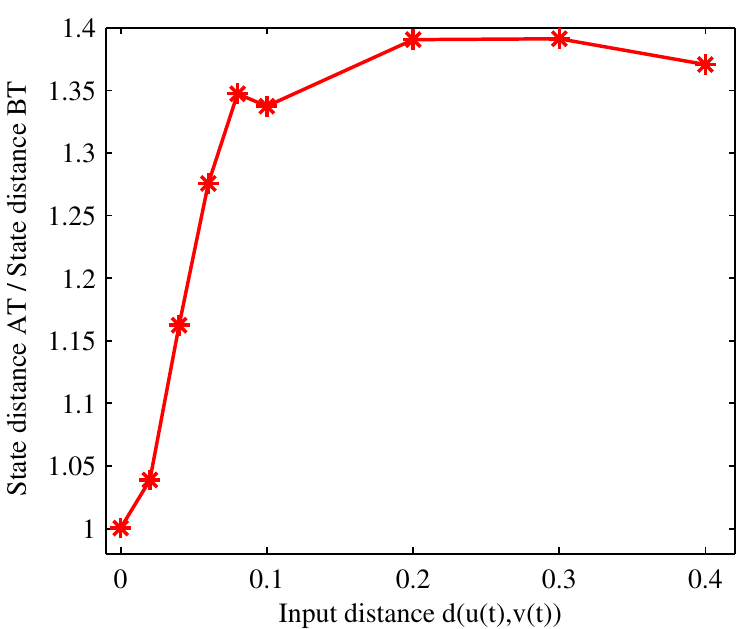}}
	\caption{(a) The internal state distances (averaged over 200 trials) of randomly generated liquids and the same liquids when trained through structural plasticity is plotted against the distances between the input spike trains. While at lower input distances (intra-class) the separation achieved by both are similar, the trained liquid provides more separation at higher input distances (inter-class). (b) The ratio (averaged over 200 trials) of state distance obtained by trained and random liquids gradually increases with input distance and saturates approximately at 1.36 $\pm$ 0.18.}
	\label{d_out_vs_d_in}
\end{figure}

Moreover, to evaluate the separability across a wide range of input distances, we generate numerous spike train pairs having progressively increasing distances and note the internal state distances of both the random and trained liquids. While the state distances at $t = T_p = 1 \; sec$ for both the random and trained liquids are plotted in Fig. \ref{d_out_vs_d_in}(a), their ratio is shown in Fig. \ref{d_out_vs_d_in}(b). Fig. \ref{d_out_vs_d_in}(a) and Fig. \ref{d_out_vs_d_in}(b) suggest that for inputs with smaller distance, which might correspond to the intra-class separation during classification tasks, the separability provided by both the liquids are close. On the other hand, when the distance between inputs increase, which might correspond to inter-class separation for classification tasks, the separation provided by trained liquids are more than random liquid. According to Fig. \ref{d_out_vs_d_in}(b) the ratio of separation provided by our liquid trained by structural plasticity and random liquid increases and finally saturates at ~1.36. Hence, our trained liquid provides 1.36 $\pm$ 0.18 times more inter-class separation than a random liquid while maintaining a similar intra-class separation. The increased separation achieved by our morphological learning rule provides the subsequent linear classifier stage with an easier recognition problem.

The \emph{pairwise separation property} is a good measure of the extent to which the details of the input streams are captured by the liquid's internal state. However, in most real-world problems we require the liquid to not only produce a desired internal state for two, but for a fairly large number of $m$ significantly different input spike trains. Although we could test whether a liquid can separate each of the $m \choose 2$ pairs of such inputs, we still would not know whether a subsequent linear classifier would be able to generate given target outputs for these $m$ inputs. Hence, a stronger measure of kernel quality is required. \cite{Maass05} addressed this issue and proposed a rank-based measure termed as the \emph{linear separation property} as a more suitable metric for evaluating a liquid's kernel quality or computational power. Although for the sake of completeness we discuss here how this quantitative measure is calculated for $m$ given spike trains, we invite the reader to look into \citep{Maass05} for its detailed proof. 

\begin{figure}[!htbp]
   \begin{center}
	\includegraphics[width=0.95\textwidth]{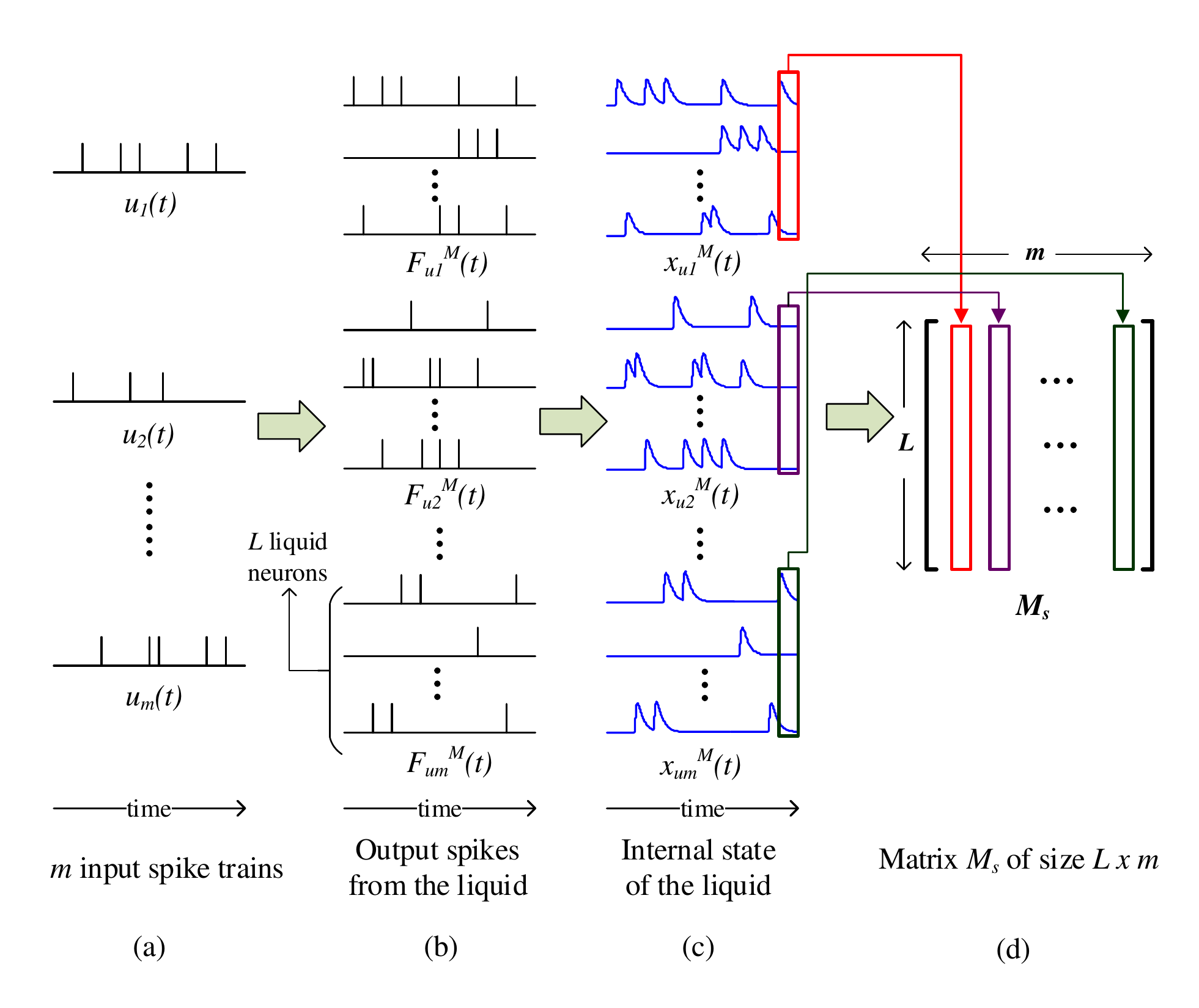}
	\caption{In this figure we explain the process of creating the matrix $M_s$, the rank of which is used to assess the quality of a liquid. (a) $m$ different input spike trains of duration $T_p$ are shown. (b) The input spike trains are injected into the liquid having $L$ neurons to obtain the post-synaptic spikes from all the neurons. $F_{u_i}^M(t)$ denotes the firing profile of the liquid when the $i^{th}$ input spike train is presented. (c) $F_{u_i}^M(t)$ is passed through an exponential kernel to generate the liquid's internal states $x^M_{u_i}(t)$. (d) The internal states at the end of the spike trains i.e. $x^M_{u_i}(T_p)$ are read for each input spike train, to create the matrix $M_s$.    }
	\label{rank_measure}
	\end{center}
\end{figure}

\begin{figure} [!htbp]
	\centering 
	\includegraphics[width=0.7\textwidth]{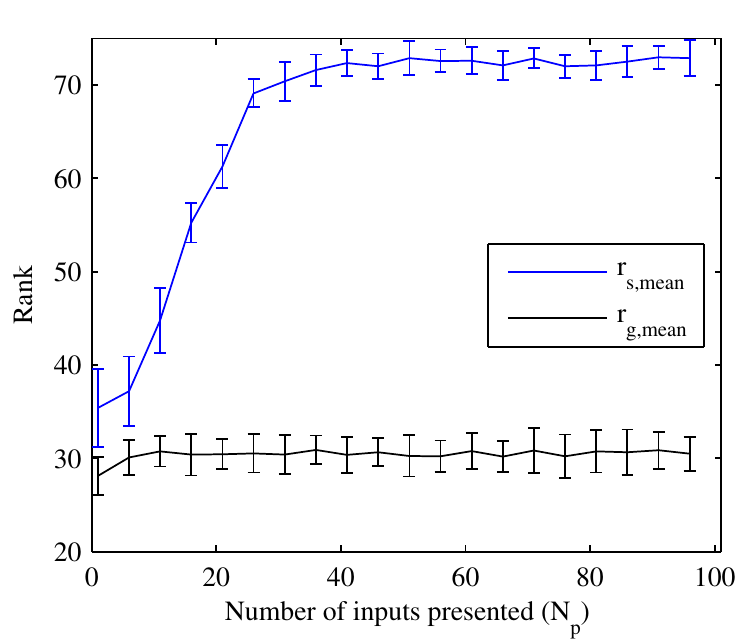}
	\caption{(Caption in the following page.)}
	\label{sep_gen_comb}
\end{figure}

The method for evaluating the \emph{linear separation property} for a liquid C for $m$ different spike trains is shown in Fig. \ref{rank_measure}. First, these spike trains are injected into the liquid and the internal its states are noted (Fig. \ref{rank_measure}(a)-(c)). Next, a $L \times m$ matrix $M_s$ is formed (Fig. \ref{rank_measure}(d)) for all the inputs $u_1(t), u_2(t),..,u_i(t),..., u_m(t)$, whose columns are the liquid states $x_{u_i}^M(T_p)$ resulting at the end of the preceding input spike train $u_i$ of duration $T_p$. \cite{Maass05} suggested that the rank $r_s$ of matrix $M_s$ reflects the \emph{linear separation property} and can be considered as a measure of the kernel quality or computational power of the liquid $C$, since $r_s$ is the `number of degrees of freedom' that a subsequent linear classifier has in assigning target outputs $y_i$ to these inputs $u_i$. Hence, a liquid with more computational power or better kernel quality has a higher value of $r_s$. The proposed unsupervised learning rule modifies the liquid connections after the end of each pattern. Hence in real-time simulations, where patterns are presented continuously, the connections within our liquid are modified through structural plasticity in an online fashion. To show that the online update generates liquids with progressively better kernel quality, we plot in Fig. \ref{sep_gen_comb} the value of the mean rank $r_{s,mean}$ against the number of spike trains ($N_p$) applied to the circuit. $r_{s,mean}$ is calculated by taking an average of the ranks $r_s$ obtained from each trial. In a real world implementation the liquid connections will get updated without any intervention and hence calculation of rank is not required. However, to probe into the performance of our learning rule, we create the matrix $M_s$ after each input is presented based on all the input spike trains at our disposal and calculate its rank $r_s$. The curves presented in Fig. \ref{sep_gen_comb} are generated by applying 100 different spike trains to liquids having random initial states for each trial. Each point in the curves is averaged over 200 trials. Fig. \ref{sep_gen_comb} clearly shows that the average rank $r_{s,mean}$ increases as more number of inputs are presented until it reaches saturation. This suggests that our proposed unsupervised structural plasticity based learning rule is capable of generating liquids with more computational power as compared to the traditional method of randomly generating the liquid connections. Note that the point in the curve $r_{s,mean}=35.48$ corresponding to $N_p=1$ reflects the kernel quality or computational power of the traditional randomly generated liquid. The proposed algorithm is able to generate liquids having $r_{s,mean}=72.87$ i.e. 2.05 $\pm$ 0.27 times better than random liquids.  

\addtocounter{figure}{-1}
\begin{figure} [t!]
	\caption{This figure shows the rank of the matrices $M_s$ and $M_g$, averaged over 200 trials and denoted by $r_{s,mean}$ and $r_{g,mean}$ respectively, against the number of input spike trains $N_p$ presented in succession. At each point of the curves, the rank is calculated based on all the spike trains. Hence, the value of $r_{s,mean}$ and $r_{g,mean}$ for $N_p=1$ corresponds to a randomly generated liquid. Our learning rule trains this randomly generated liquid through structural plasticity as the input spike trains are gradually presented. The curve shows that $r_{s,mean}$ gradually increases until it reaches saturation at $N_p=41$. Moreover, this figure depicts that by applying an unsupervised low-resolution training mechanism, we are able to generate liquids with 2.05 $\pm$ 0.27 times more computational power or better kernel quality as compared to traditional randomly liquids. Furthermore, this figure throws some light on the generalization performance of liquids trained through our proposed structural plasticity based learning rule. The matrix $M_g$ is formed from liquids produced at each stage of training corresponding to the evolution of $r_{s,mean}$. Its rank (averaged over 200 trials) denoted by $r_{g,mean}$ is plotted against the number of different input spike trains $N_p$ presented successively presented to the liquid. The almost flat line represents that our learning rule can retain the generalization ability of random liquids. At a particular value of $N_p$, $r_{s,mean}$ and $r_{g,mean}$ are calculated based on the same liquid thereby providing an insight to both its separation and generalization property. \\}
\end{figure}

\subsection{Generalization capability}
Till now we have looked into the separation capability of a liquid to assess its computational performance. However, this is only one piece of the puzzle with another being its capability to \emph{generalize} a learned computational function to new inputs i.e. input it has not seen during the training phase. It is interesting to note that \cite{Maass05} suggested using the same rank measure used in Sec. \ref{sep_cap} to measure a liquid's generalization capability. However, in this case, the inputs to the liquid are entirely different from the ones used in Sec. \ref{sep_cap}. For a particular trial, a single spike train is considered and $s$ noisy variations of it are created to form the set $S_{univ}$. In other words, $S_{univ} = \{ u_1(t), u_2(t), ......, u_s(t)\}$ contains many jittered versions of the same input signal. Similar to Sec. \ref{sep_cap} a $L\times s$ matrix $M_{g}$ is formed by injecting these spike trains into a liquid and by noting the liquid states resulting at the end of the preceding input spike train $u_i$ of duration $T_p$. However, unlike Sec. \ref{sep_cap} a lower value of rank $r_{g}$ of the matrix $M_{g}$ corresponds to better generalization performance. To assess our liquid's generalization capability we provide $s=100$ input spike trains to the liquids produced at each stage of learning corresponding to the evolution of $r_{s,mean}$ curve in Fig. \ref{sep_gen_comb} and note the average rank $r_{g,mean}$ of matrix $M_{g}$. $r_{g,mean}$. $r_{g,mean}$ vs. the number inputs ($N_p$) presented to the liquid is shown in Fig. \ref{sep_gen_comb}. An almost flat curve shown in Fig. \ref{sep_gen_comb} suggests that our morphologically trained liquids are capable of retaining the generalization performance shown by random liquids. This revelation combined with the insight from Sec. \ref{sep_cap} suggests that our trained liquids are capable of amplifying the inter-class separation while retaining the intra-class distances for classification problems.   

\subsection{Generality}  
\begin{figure}[!t]
	\begin{center}
		\includegraphics[width=0.7\textwidth]{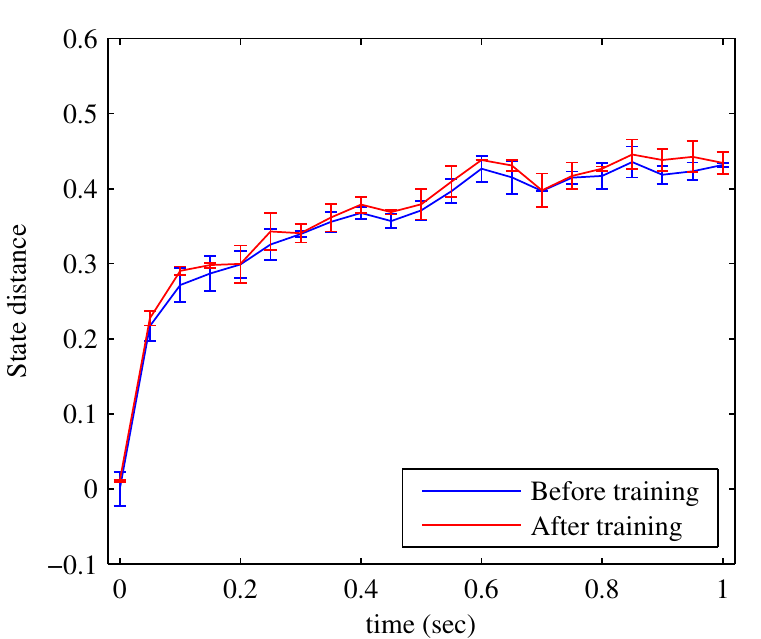}
		\caption{In this figure, we assess the effect of training on the generality of liquids. The blue and the red curves (averaged over 200 trials) correspond to the separation of random and trained liquids respectively when they are injected with 500 previously unseen and significantly different (with respect to the training set) input spike train pairs of distance $d(u(t),v(t)) = 0.2$. The comparable curves obtained by both the random and trained liquids suggest that our structural plasticity rule does not decrease a liquid's generality.}
		\label{generality}
	\end{center}
\end{figure}

By performing the previous experiments we got a fair idea about the separation and generalization property of our trained liquids. Next, we look into its generality i.e. whether the trained liquid is still general enough to separate inputs which it has not seen before and which are not related (noisy, jittered, etc.) to the training inputs in any way. We consider the initial random liquid before training and the final trained liquid after training on 100 spike train inputs of Fig. \ref{sep_gen_comb} from each trial and separately inject both of them with 500 different spike train pairs $u_i(t)$ and $v_i(t)$ having $d(u(t),v(t)) = 0.2$. We note the average state distances of both the liquids for these 500 spike train pairs for the current trial. This experiment is repeated for each trial of Fig. \ref{sep_gen_comb} and the mean state distances are shown in Fig. \ref{generality}. It is clear from the two closely placed curves of Fig. \ref{generality} that the trained and random liquids show similar generality. We conclude that even if a liquid is trained on a set of inputs, it is still fairly general to previously unseen significantly different inputs.

\subsection{Fading memory}
Till now we discussed the \emph{separation} property, \emph{generalization} capability and generality of liquids. Another component which defines a liquid is its \emph{fading memory}. Since a liquid is a recurrent interconnection of spiking neurons, the effect of an input applied to it may be felt at its output even after it is gone. A liquid with a superior fading memory is able to remember a given input activity for a longer duration of time. To study the effect of our structural plasticity rule on a liquid's fading memory we performed the following experiment. First, an input spike train of duration $T_p=1 sec$ was generated having a burst of spikes at a random time. Next, this spike train was injected into a randomly generated liquid and the post-synaptic spikes of its neurons were recorded. Subsequently, this liquid was trained through our learning rule and the same spike train was reapplied. This experiment was repeated N times for different spike trains and liquids and time of the last spike at the liquid output were noted for all these trials to compute the following measure:
\begin{equation}
	TLS_{diff} = \frac{1}{N} \sum\limits_{i=1}^{N} \left( t_i^{tr,last} -  t_i^{ran,last} \right)
\end{equation}   
where $t_i^{ran,last}$ and $t_i^{tr,last}$ are the time to last spike at the output of random and trained liquids for the $i^{th}$ trial. The computed metric $TLS_{diff}$ serves as a parameter to assess the amount of additional fading memory provided by our training mechanism. For our simulations we keep N=200.

\begin{figure}[!htbp]
	\begin{center}
		\includegraphics[width=0.95\textwidth]{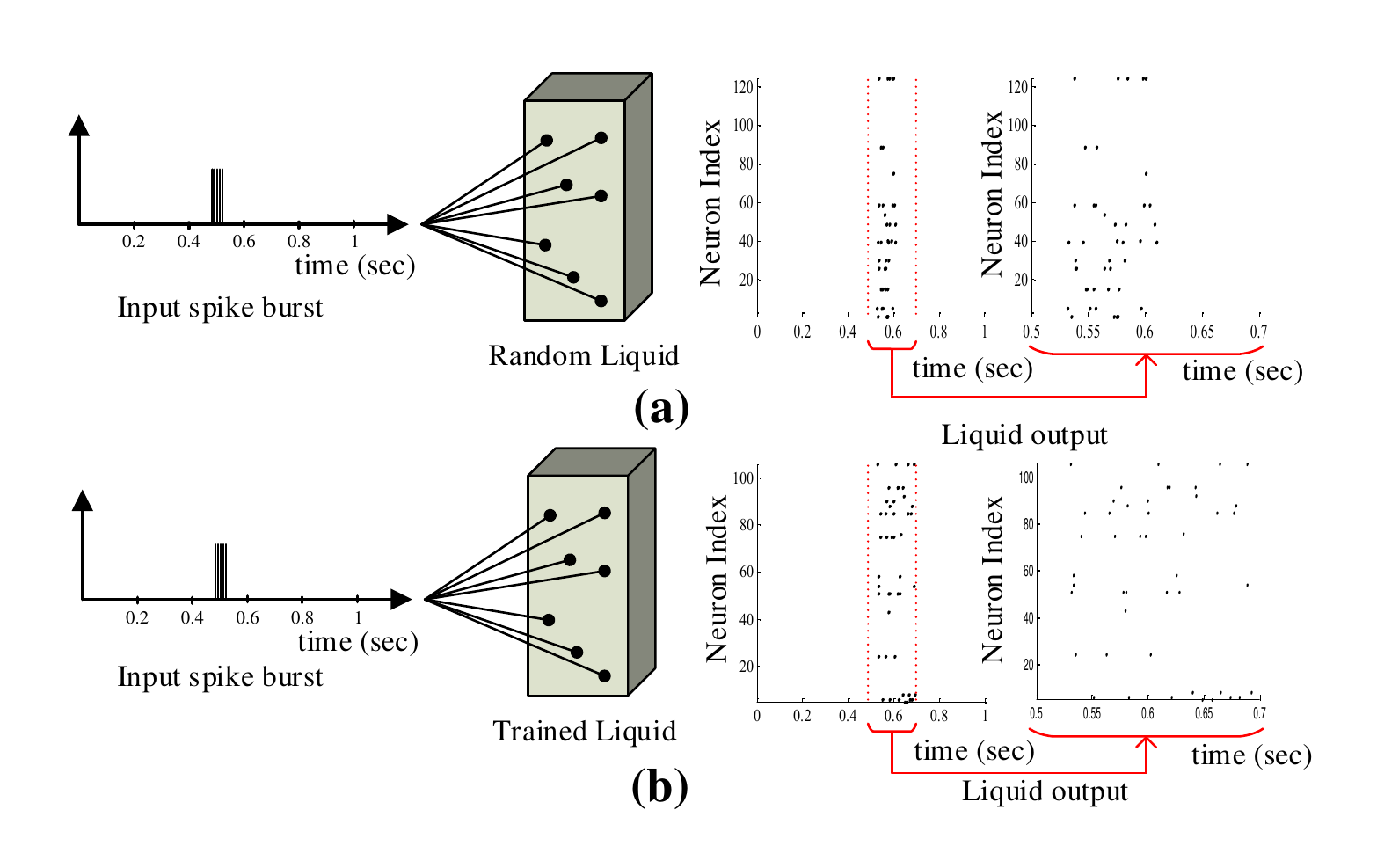}
		\caption{(a) An input spike train with a burst of spikes around 500 $ms$ is presented to a randomly generated liquid. Its output is shown and the portion of its output where spikes are present is magnified. (b) The random liquid is taken and trained by our structural plasticity rule. Subsequently, it is injected with the same input spike train and its output (along with the magnified version) is shown. Comparing the last two figures of (a) and (b) it is clear that our learning rule endows the liquid with a higher fading memory.}
		\label{memory_comparison}
	\end{center}
\end{figure}

In Fig. \ref{memory_comparison}, we show the outcome of a single trial. Fig. \ref{memory_comparison}(a) shows that an input spike train with a burst of spikes around 500 $ms$ is given as input to a random liquid and its corresponding output is recorded. We train this random liquid through the proposed algorithm and show in Fig. \ref{memory_comparison}(b) its output when the same spike train is reapplied. It can be seen from Fig. \ref{memory_comparison} that $t_i^{ran,last}=$ 0.6128 $secs$ and $t_i^{tr,last}=$ 0.6929 $secs$ i.e. the last spike provided by the trained liquid occurs 80.1 $ms$ after the one provided by the random liquid. Combining the result from all the trials we obtain that trained liquids provide $TLS_{diff}=$ 83.67 $\pm$ 5.79 $ms$ longer fading memory than random liquids having 92.8 $\pm$ 5.03 $ms$ fading memory. This experiment suggests that the fading memory of a liquid can be increased by applying structural plasticity. 

\subsection{Liquid connectivity}  
\begin{figure}[!htbp]
	\begin{center}
		\includegraphics[width=0.95\textwidth]{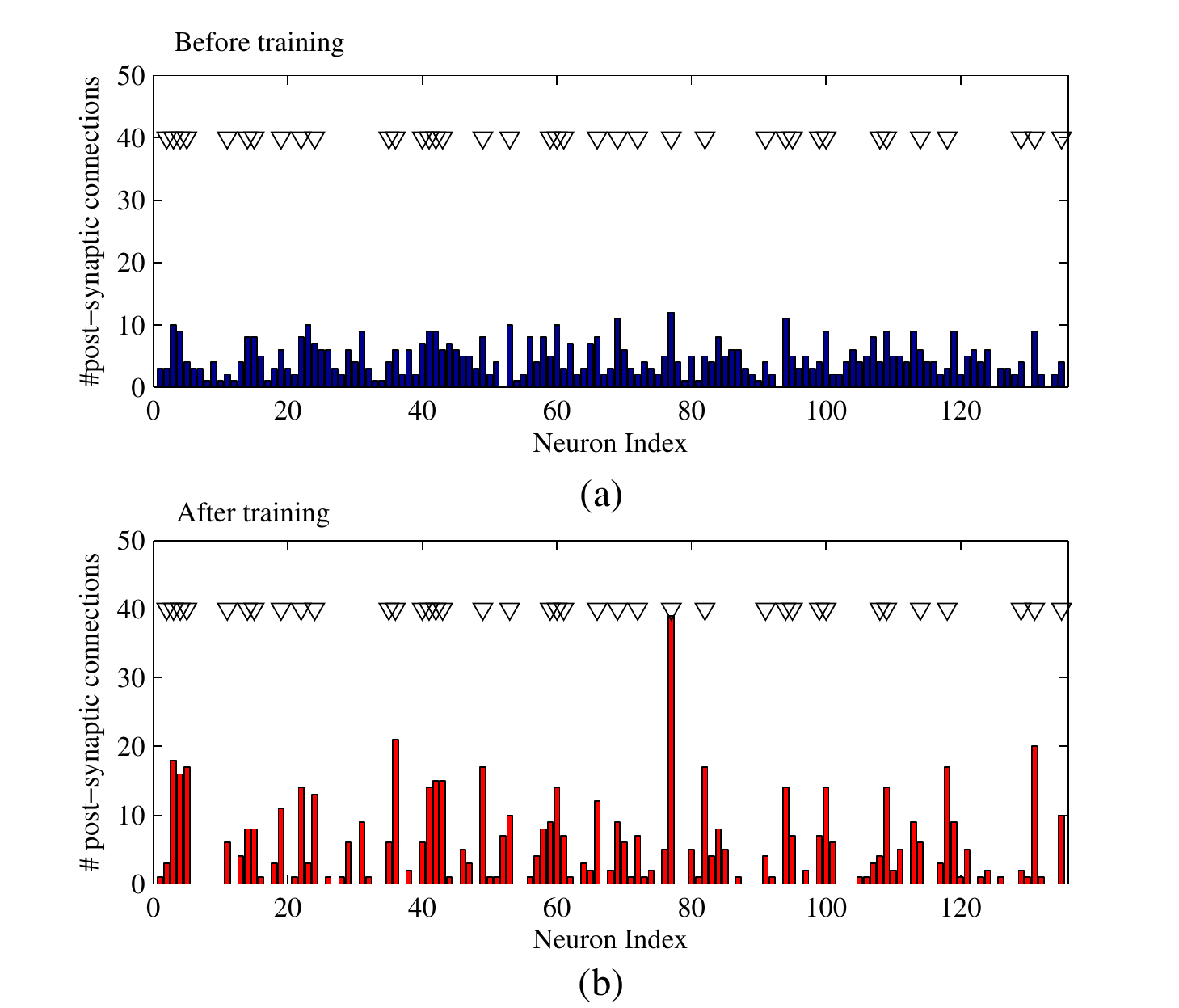}
		\caption{(a) The number of post-synaptic connections vs. the neuron index is shown before training i.e. for the random liquid. The connections are uniformly distributed across the neurons of the liquid. (b) The number of post-synaptic connections vs. the neuron index is shown after the training is complete. The connections get rearranged during learning in such a way that after training some neurons have more post-synaptic connections than the others. Note that the total number of connections are same for both the figures since it remains constant throughout the learning procedure. Our learning rule does not create any new connections, instead, it reorganizes them. The downward-triangle denotes the neurons which have the input line as pre-synaptic connection. The figure reveals that most of the neurons having increased post-synaptic connections have the input line as a pre-synaptic connection. }
		\label{connectivity_update}
	\end{center}
\end{figure}

In the previous sections, we have analyzed the output of our morphologically trained liquids in various ways and for different inputs. Here we will delve into the liquid itself, and analyze the effect of our learning rule on the recurrent connectivity. Fig. \ref{connectivity_update} shows a representative example of the conducted experiments which depicts the number of post-synaptic connections each neuron in the liquid has before (Fig. \ref{connectivity_update}(a)) and after (Fig. \ref{connectivity_update}(b)) training on $N_p$=100 distinct spike trains. The triangles in Fig. \ref{connectivity_update} identifies the neurons which have the input spike train as a pre-synaptic input. While the number of post-synaptic connections are distributed uniformly across the neurons of the random liquid as shown in Fig. \ref{connectivity_update}(a), in Fig. \ref{connectivity_update}(b) we see that some neurons have more post-synaptic connections compared to others. Since the number of connections are same for both Fig. \ref{connectivity_update}(a) and Fig. \ref{connectivity_update}(b), during learning the post-synaptic connections of some neurons increase while the others decrease. 

A close inspection of Fig. \ref{connectivity_update}(b) reveals that most of the neurons which have more post-synaptic connections than the others have the input line as a pre-synaptic connection. This essentially means that the neurons to which the input gets randomly distributed are more likely to be selected as a replacement during the connection swapping procedure (Sec. \ref{unsup_str_plas}) of our learning.    

\subsection{Comparison with other works}\label{SDSM_comp_sec} 
After its inception in \citep{Maass2002}, researchers have looked into improving the reservoir of Liquid State Machine and we have provided a brief survey of the same in Sec. \ref{research_liquid}. Out of the algorithms showcased in Sec. \ref{research_liquid}, we compare the proposed learning rule with the work, termed as SDSM, provided in \citep{Norton2010} since it is architecturally similar to our algorithm and iteratively updates a randomly generated liquid like us. The authors of \citep{Norton2010} derived a metric based on the separation property of the liquid and used it to update the synaptic parameters. We consider the pattern recognition task described in \citep{Norton2010} and compare the performance of the proposed algorithm with the results obtained by SDSM \citep{Norton2010}. 

In this task, a dataset with variable number of classes of spiking patterns is considered. Each pattern has eight input dimensions and patterns of each class are generated by creating jittered versions of a random template. The random template for each class is generated by plotting individual spikes with a random distance between one another. This distance is drawn from the absolute value of a normal distribution with a mean of 10 $ms$ and a standard deviation of 20 $ms$. The amount of jitter added to each spike is randomly drawn from a normal distribution with zero mean and 5 $ms$ standard deviation. Similar to \citep{Norton2010}, we consider four, eight and twelve class versions of this problem. The number of training and testing patterns per class have been kept to 400 and 100 respectively.

For fair comparison, the readout and the experimental setup have been kept similar to \citep{Norton2010}. The readout is implemented by a single layer of perceptrons where each perceptron is trained to identify a particular class of patterns. The results are averaged over 50 trials and in each trial the liquid is evolved for 500 iterations. The liquids are constructed with LIF neurons and static synapses and the parameters have been set to the values listed in \citep{Norton2010}.

\begin{figure} [!htbp]
	\centering 
	\subfloat[]{\includegraphics[width=0.7\textwidth]{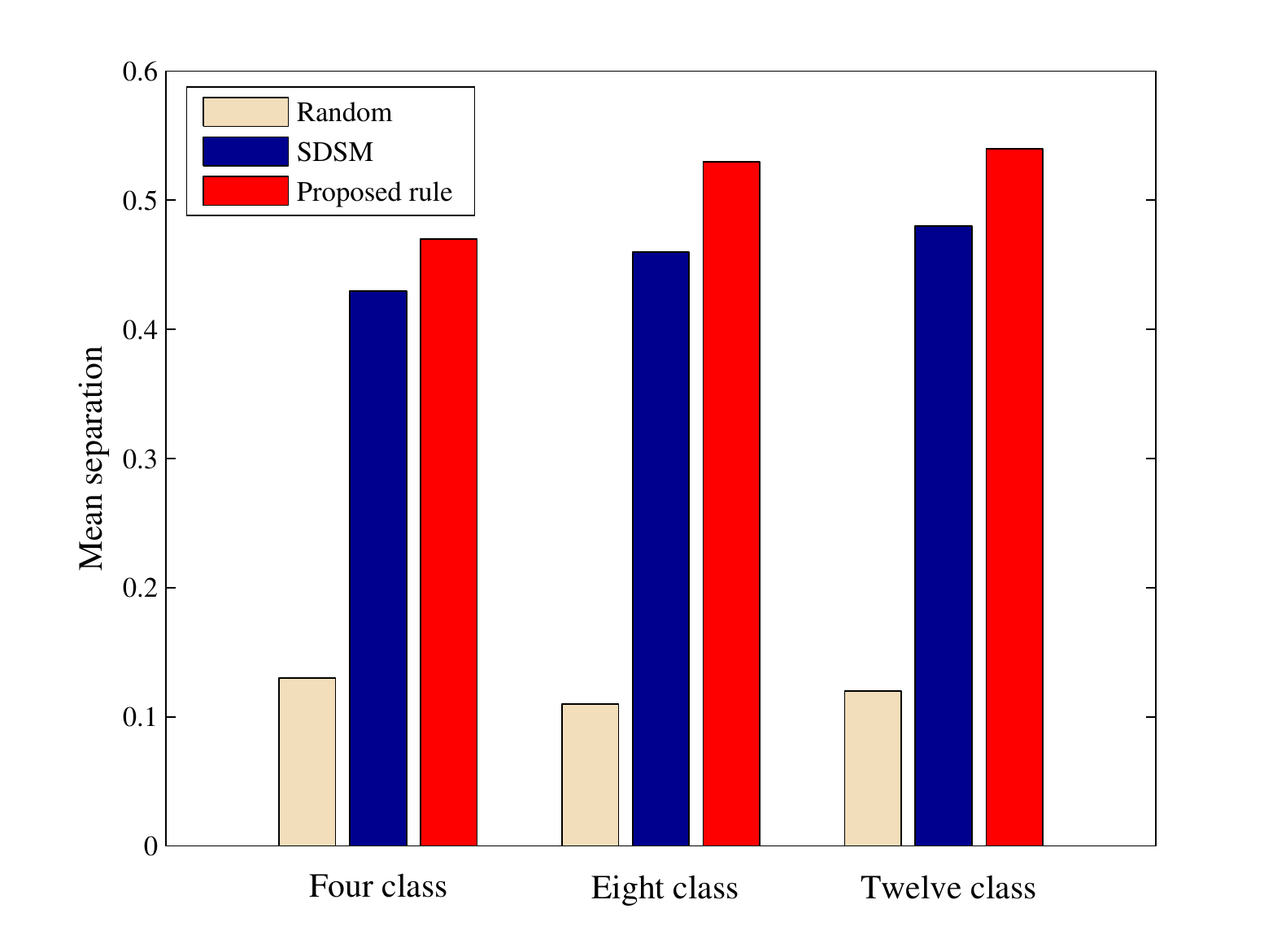}}\\
	\subfloat[]{\includegraphics[width=0.7\textwidth]{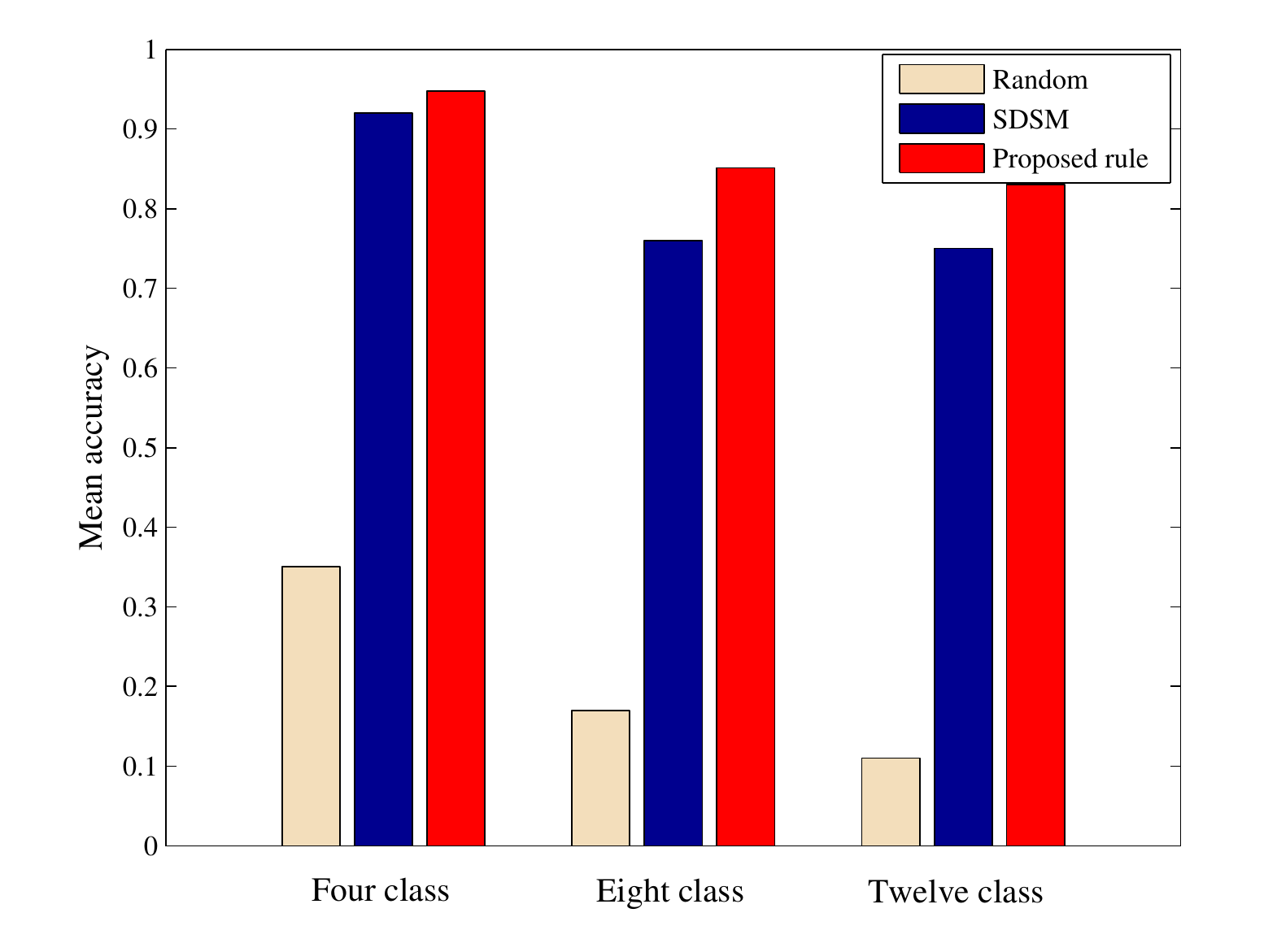}}
	\caption{(a) Mean separation and (b) mean accuracy are shown for random liquids and for the same random liquid when trained separately through SDSM and the proposed structural plasticity based learning rule. The results depict that our algorithm outperforms the SDSM learning rule.}
	\label{SDSM_comp}
\end{figure}

The liquid separation and classification accuracy for the testing patterns, averaged over 50 trials, are plotted in Fig. \ref{SDSM_comp}(a) and Fig. \ref{SDSM_comp}(b) respectively. It is evident from Fig. \ref{SDSM_comp} that LSMs with liquid evolved through structural plasticity obtain superior performance as compared to LSMs trained through SDSM and LSMs with random liquid. Quantitatively, our algorithm provides  9.30\%, 15.21\% and 12.52\% increase in liquid separabilities and 2.8\%, 9.1\% and 7.9\% increase in classification accuracies for four, eight and twelve class recognition tasks respectively, as compared to SDSM. Since the proposed learning rule creates liquids with higher separation as shown in Fig. \ref{SDSM_comp}(b), the readout is able to provide better classification accuracies.  
   
\section{Conclusion}
In this article, we have proposed an unsupervised learning rule which trains the liquid or reservoir of LSM by rearranging the synaptic connections. The proposed learning rule does not modify synaptic weights and hence keeps the average synaptic weight of the liquid constant throughout learning. Since it only involves modification and storage of the connection matrix during learning, it can be easily implemented by AER protocols. An analysis of the `pairwise separation property' reveals that liquids trained through the proposed learning rule provide a 1.36 $\pm$ 0.18 times more inter-class separation while maintaining similar intra-class separation as compared to the traditional random liquid. Next we looked into the `linear separation property' and from the performed experiments it is clear that our trained liquids are 2.05 $\pm$ 0.27 times better than random liquids. Moreover, experiments performed to test the `generalization property' and `generality' of liquids formed by our learning algorithm reveal that they are capable of inheriting the performance provided by random liquids. Furthermore, we have shown that our trained liquids have 83.67 $\pm$ 5.79 $ms$ longer fading memory than random liquids providing 92.8 $\pm$ 5.03 $ms$ fading memory for a particular type of spike train inputs. These results suggest that our learning rule is capable of eliminating the weaknesses of a random liquid while retaining its strengths. We have also analyzed the evolution of internal connections of the liquid during training. Furthermore, we have shown that compared to a recently proposed method of liquid evolution termed as SDSM, we provide 9.30\%, 15.21\% and 12.52\% more liquid separations and 2.8\%, 9.1\% and 7.9\% better classification accuracies for four, eight and twelve class classifications respectively on a task described in Sec. \ref{SDSM_comp_sec}.       

The plans for our future work include developing a framework that combines the proposed liquid with the readout proposed in \citep{roy_tbcas} which is composed of neurons with nonlinear dendrites and binary synapses and trained through structural plasticity. \citep{amitava_iscas_2015} have proposed a hardware implementation of this readout and the next step is to implement the proposed liquid in hardware. Subsequently, we will combine them to form a complete structural plasticity based LSM system and deploy it for real-time applications. Moreover, having achieved success in applying structural plasticity rules to train a generic spiking neural recurrent architecture, we will move forward to develop spike-based morphological learning rules for multi-layer feedforward spiking neural networks. We will employ these networks to classify individuated finger and wrist movements of monkeys \citep{NITISH1} and recognize spoken words \citep{Verstraeten}.  

\section*{Appendix}
In this table, we list specification of the liquid architecture and values of the parameters used in this paper. Unless otherwise mentioned, these are the values used for the experiments.
\begin{longtable}{|p{.65\textwidth}|p{.3\textwidth}|}
			\hline
			 \multicolumn{2}{|c|}{\textbf{Liquid specification}} \\
			 \hline Number of neurons ($L$) & 135\\
			 \hline Percentage of excitatory neurons ($\frac{L_e*100}{L}$) & 80\\
			 \hline Percentage of inhibitory neurons ($\frac{L_i*100}{L}$) & 20\\
			 \hline Structure & Single $15\times3\times3$ column\\
			 \hline Excitatory-excitatory connection probability & 0.3\\
			 \hline Excitatory-inhibitory connection probability & 0.2\\
			 \hline Inhibitory-excitatory connection probability & 0.4\\
			 \hline Inhibitory-inhibitory connection probability & 0.1\\
			 \hline
			 \hline \multicolumn{2}{|c|}{\textbf{Leaky Integrate and Fire (LIF) neuron parameters}} \\
			 \hline Membrane time constant & 30 $ms$\\
			 \hline Input resistance & 1 $M\Omega$\\
			 \hline Absolute refractory period of excitatory neurons & 3 $ms$\\
			 \hline Absolute refractory period of inhibitory neurons & 2 $ms$\\
			 \hline Threshold voltage & 15 $mV$\\
			 \hline Reset voltage & 13.5 $mV$\\
			 \hline Constant nonspecific background current & 13.5 $nA$\\
			 \hline
			 \hline \multicolumn{2}{|c|}{\textbf{Dynamic Synapse parameters}} \\
			 \hline Excitatory-excitatory $U_{mean}$ & 0.5\\
			 \hline Excitatory-excitatory $D_{mean}$ & 1.1 $sec$\\
			 \hline Excitatory-excitatory $F_{mean}$ & 0.05 $sec$\\
			 \hline Excitatory-inhibitory $U_{mean}$ & 0.05\\
			 \hline Excitatory-inhibitory $D_{mean}$ & 0.125 $sec$\\
			 \hline Excitatory-inhibitory $F_{mean}$ & 1.2 $sec$\\
			 \hline Inhibitory-excitatory $U_{mean}$ & 0.25\\
			 \hline Inhibitory-excitatory $D_{mean}$ & 0.7 $sec$\\
			 \hline Inhibitory-excitatory $F_{mean}$ & 0.02 $sec$\\
			 \hline Inhibitory-inhibitory $U_{mean}$ & 0.32\\
			 \hline Inhibitory-inhibitory $D_{mean}$ & 0.144 $sec$\\
			 \hline Inhibitory-inhibitory $F_{mean}$ & 0.06 $sec$\\
			 \hline Standard deviation of $U$, $D$ and $F$ & 50\% of respective mean\\
			 \hline Time constant of excitatory synapse ($\tau_s$) & 3 $ms$\\ 
			 \hline Time constant of inhibitory synapse & 6 $ms$\\
			 \hline Excitatory-excitatory transmission delay & 1.5 $ms$\\
			 \hline Transmission delay of other connections & 0.8 $ms$\\ 
			 \hline	
			 \hline \multicolumn{2}{|c|}{\textbf{Structural Plasticity parameters}} \\
			 \hline Number of silent synapses ($n_R$) & 25\\	 	
			 \hline	Slow time constant of kernel $K$ ($\tau_s$) & 3 $ms$\\
			 \hline	Fast time constant of kernel $K$ ($\tau_f$) & Very small positive value\\
			 \hline	 			 
\caption{Liquid specification and parameter values}
	\label{params_table}
\end{longtable}
		
\subsection*{Acknowledgments}
Financial support from MOE through grant ARC 8/13 is acknowledged.

\bibliographystyle{chicago}

\begin{thebibliography}{}
	
	\bibitem[\protect\citeauthoryear{Aggarwal, Acharya, Tenore, Shin, Cummings,
		Schieber, and Thakor}{Aggarwal et~al.}{2008}]{NITISH1}
	Aggarwal, V., S.~Acharya, F.~Tenore, H.~Shin, R.~E. Cummings, M.~Schieber, and
	N.~Thakor (2008, Feb.).
	\newblock Asynchronous decoding of dexterous finger movements using m1 neurons.
	\newblock {\em IEEE Transactions on Neural Systems and Rehabilitation
		Engineering\/}~{\em 16\/}(1), 3--14.
	
	\bibitem[\protect\citeauthoryear{Arthur and Boahen}{Arthur and
		Boahen}{2006}]{Arthur2006}
	Arthur, J.~V. and K.~Boahen (2006).
	\newblock Learning in silicon: Timing is everything.
	\newblock In ong (Ed.), {\em Advances in Neural Information Processing
		Systems}.
	
	\bibitem[\protect\citeauthoryear{Banerjee, Bhaduri, Roy, Kar, and
		Basu}{Banerjee et~al.}{2015}]{amitava_iscas_2015}
	Banerjee, A., A.~Bhaduri, S.~Roy, S.~Kar, and A.~Basu (2015).
	\newblock A current-mode spiking neural classifier with lumped dendritic
	nonlinearity.
	\newblock In {\em Proceedings of the IEEE International Sympoisum on Circuits
		and Systems (ISCAS)}, Number May.
	
	\bibitem[\protect\citeauthoryear{Brader, Senn, and Fusi}{Brader
		et~al.}{2007}]{Brader2007}
	Brader, J., W.~Senn, and S.~Fusi (2007, May).
	\newblock Learning real-world stimuli in a neural network with spike-driven
	synaptic dynamics.
	\newblock {\em Neural Computation\/}~{\em 19\/}(11), 2881--2912.
	
	\bibitem[\protect\citeauthoryear{Florian}{Florian}{2013}]{Chronotron}
	Florian, R.~V. (2013).
	\newblock The chronotron: a neuron that learns to fire temporally precise spike
	patterns.
	\newblock {\em PLoS ONE\/}~{\em 7\/}(8), e40233.
	
	\bibitem[\protect\citeauthoryear{Frid, Hazan, and Manevitz}{Frid
		et~al.}{2012}]{Frid2012}
	Frid, A., H.~Hazan, and L.~Manevitz (2012, Nov.).
	\newblock Temporal pattern recognition via temporal networks of temporal
	neurons.
	\newblock In {\em Proceedings of the 27th Convention of Electrical Electronics
		Engineers in Israel (IEEEI)}, pp.\  1--4.
	
	\bibitem[\protect\citeauthoryear{Gardner, Sporea, and Gr\"{u}ning}{Gardner
		et~al.}{2015}]{gardner2015}
	Gardner, B., I.~Sporea, and A.~Gr\"{u}ning (2015).
	\newblock Learning spatiotemporally encoded pattern transformations in
	structured spiking neural networks.
	\newblock {\em Neural Computation\/}.
	
	\bibitem[\protect\citeauthoryear{George, Diehl, Cook, Mayr, and
		Indiveri}{George et~al.}{2015}]{George2015b}
	George, R., P.~Diehl, M.~Cook, C.~Mayr, and G.~Indiveri (2015, July).
	\newblock Modeling the interplay between structural plasticity and
	spike-timing-dependent plasticity.
	\newblock {\em BMC Neuroscience\/}~{\em 16\/}(Suppl 1), P107.
	
	\bibitem[\protect\citeauthoryear{George, Mayr, Indiveri, and Vassanelli}{George
		et~al.}{2015}]{George2015a}
	George, R., C.~Mayr, G.~Indiveri, and S.~Vassanelli (2015, June).
	\newblock Event-based softcore processor in a biohybrid setup applied to
	structural plasticity.
	\newblock In {\em Event-based Control, Communication, and Signal Processing
		(EBCCSP), 2015 International Conference on}, pp.\  1--4.
	
	\bibitem[\protect\citeauthoryear{Gerstner and Kistler}{Gerstner and
		Kistler}{2002}]{Gerstner2002}
	Gerstner, W. and W.~Kistler (2002).
	\newblock {\em Spiking Neuron Models: An Introduction}.
	\newblock New York, NY, USA: Cambridge University Press.
	
	\bibitem[\protect\citeauthoryear{Gutig and Sompolinsky}{Gutig and
		Sompolinsky}{2006}]{Gutig2006}
	Gutig, S. and H.~Sompolinsky (2006, Feb.).
	\newblock The tempotron: a neuron that learns spike timing-based decisions.
	\newblock {\em Nature Neuroscience\/}~{\em 9\/}(1), 420--428.
	
	\bibitem[\protect\citeauthoryear{Hazan and Manevitz}{Hazan and
		Manevitz}{2012}]{Hazan2012}
	Hazan, H. and L.~M. Manevitz (2012, Feb.).
	\newblock Topological constraints and robustness in liquid state machines.
	\newblock {\em Expert Syst. Appl.\/}~{\em 39\/}(2), 1597--1606.
	
	\bibitem[\protect\citeauthoryear{Hempel, Hartman, Wang, Turrigiano, and
		Nelson}{Hempel et~al.}{2000}]{Hempel2000}
	Hempel, C.~M., K.~H. Hartman, X.~J. Wang, G.~G. Turrigiano, and S.~B. Nelson
	(2000).
	\newblock Multiple forms of short-term plasticity at excitatory synapses in rat
	medial prefrontal cortex.
	\newblock {\em Journal of Neurophysiology\/}~{\em 83\/}(5), 3031--3041.
	
	\bibitem[\protect\citeauthoryear{Hourdakis and Trahanias}{Hourdakis and
		Trahanias}{2013}]{Hourdakis2013}
	Hourdakis, E. and P.~Trahanias (2013, May).
	\newblock Use of the separation property to derive liquid state machines with
	enhanced classification performance.
	\newblock {\em Neurocomputing\/}~{\em 107}, 40--48.
	
	\bibitem[\protect\citeauthoryear{Hussain, Liu, and Basu}{Hussain
		et~al.}{2015}]{Hussain2014_nc}
	Hussain, S., S.~C. Liu, and A.~Basu (2015).
	\newblock Hardware-amenable structural learning for spike-based pattern
	classification using a simple model of active dendrites.
	\newblock {\em Neural Computation\/}~{\em 27\/}(4), 845--897.
	
	\bibitem[\protect\citeauthoryear{Ju, Xu, Chong, and Vandongen}{Ju
		et~al.}{2013}]{Ju2013}
	Ju, H., J.~X. Xu, E.~Chong, and A.~M.~J. Vandongen (2013, Feb.).
	\newblock Effects of synaptic connectivity on liquid state machine performance.
	\newblock {\em Neural Networks\/}~{\em 38}, 39--51.
	
	\bibitem[\protect\citeauthoryear{Kello and Mayberry}{Kello and
		Mayberry}{2010}]{Kello2010}
	Kello, C. and M.~Mayberry (2010, Jul.).
	\newblock Critical branching neural computation.
	\newblock In {\em Proceedings of the International Joint Conference on Neural
		Networks (IJCNN)}, pp.\  1--7.
	
	\bibitem[\protect\citeauthoryear{Kuhlmann, Hauser-Raspe, Manton, Grayden,
		Tapson, and van Schaik}{Kuhlmann et~al.}{2014}]{Kuhlmann2014}
	Kuhlmann, L., M.~Hauser-Raspe, J.~H. Manton, D.~B. Grayden, J.~Tapson, and
	A.~van Schaik (2014).
	\newblock {Approximate, Computationally Efficient Online Learning in Bayesian
		Spiking Neurons}.
	\newblock {\em Neural Computation\/}~{\em 26\/}(3), 472--496.
	
	\bibitem[\protect\citeauthoryear{Maass, Legenstein, Bertschinger, and
		Graz}{Maass et~al.}{2005}]{Maass05}
	Maass, W., R.~Legenstein, N.~Bertschinger, and T.~U. Graz (2005).
	\newblock Methods for estimating the computational power and generalization
	capability of neural microcircuits.
	\newblock In {\em Advances in Neural Information Processing Systems}, pp.\
	865--872. MIT Press.
	
	\bibitem[\protect\citeauthoryear{Maass, Natschl\"{a}ger, and Markram}{Maass
		et~al.}{2002}]{Maass2002}
	Maass, W., T.~Natschl\"{a}ger, and H.~Markram (2002, November).
	\newblock Real-time computing without stable states: a new framework for neural
	computation based on perturbations.
	\newblock {\em Neural Computation\/}~{\em 14\/}(11), 2531--2560.
	
	\bibitem[\protect\citeauthoryear{Markram and Tsodyks}{Markram and
		Tsodyks}{1996}]{Markham1996}
	Markram, H. and M.~Tsodyks (1996).
	\newblock Redistribution of synaptic efficacy between neocortical pyramidal
	neurons.
	\newblock {\em Nature 382\/}~{\em 382\/}(6594), 807--810.
	
	\bibitem[\protect\citeauthoryear{Moore}{Moore}{2002}]{Moore2002}
	Moore, S.~C. (2002).
	\newblock Back-propagation in spiking neural networks.
	\newblock Master's thesis, University of Bath.
	
	\bibitem[\protect\citeauthoryear{Norton and Ventura}{Norton and
		Ventura}{2010}]{Norton2010}
	Norton, D. and D.~Ventura (2010, Oct.).
	\newblock Improving liquid state machines through iterative refinement of the
	reservoir.
	\newblock {\em Neurocomputing\/}~{\em 73\/}(16-18), 2893--2904.
	
	\bibitem[\protect\citeauthoryear{Notley and Gruning}{Notley and
		Gruning}{2012}]{Notley2012}
	Notley, S. and A.~Gruning (2012, Jun.).
	\newblock Improved spike-timed mappings using a tri-phasic spike
	timing-dependent plasticity rule.
	\newblock In {\em Proceedings of the International Joint Conference on Neural
		Networks (IJCNN)}, pp.\  1--6.
	
	\bibitem[\protect\citeauthoryear{Obst and Riedmiller}{Obst and
		Riedmiller}{2012}]{Obst2012}
	Obst, O. and M.~Riedmiller (2012, June).
	\newblock Taming the reservoir: Feedforward training for recurrent neural
	networks.
	\newblock In {\em Neural Networks (IJCNN), The 2012 International Joint
		Conference on}, pp.\  1--7.
	
	\bibitem[\protect\citeauthoryear{Petersen, Malenka, Nicoll, and
		Hopfield}{Petersen et~al.}{1998}]{Petersen1998}
	Petersen, C. C.~H., R.~C. Malenka, R.~A. Nicoll, and J.~J. Hopfield (1998).
	\newblock All-or-none potentiation at ca3-ca1 synapses.
	\newblock {\em Proc. Natl. Acad. Sci. USA\/}~{\em 95\/}(8), 4732--4737.
	
	\bibitem[\protect\citeauthoryear{Poirazi and Mel}{Poirazi and
		Mel}{2001}]{Mel2001}
	Poirazi, P. and B.~W. Mel (2001, Mar.).
	\newblock Impact of active dendrites and structural plasticity on the memory
	capacity of neural tissue.
	\newblock {\em Neuron\/}~{\em 29, no. 3}, 779--796.
	
	\bibitem[\protect\citeauthoryear{Ponulak and Kasi\'{n}ski}{Ponulak and
		Kasi\'{n}ski}{2010}]{ReSuMe_neco}
	Ponulak, F. and A.~Kasi\'{n}ski (2010, Feb.).
	\newblock Supervised learning in spiking neural networks with resume: Sequence
	learning, classification, and spike shifting.
	\newblock {\em Neural Computation\/}~{\em 22\/}(2), 467--510.
	
	\bibitem[\protect\citeauthoryear{Rh{\'e}Aume, Grenier, and
		Boss{\'e}}{Rh{\'e}Aume et~al.}{2011}]{RheAume2011}
	Rh{\'e}Aume, F., D.~Grenier, and L.~Boss{\'e} (2011, Oct.).
	\newblock Multistate combination approaches for liquid state machine in
	supervised spatiotemporal pattern classification.
	\newblock {\em Neurocomputing\/}~{\em 74\/}(17), 2842--2851.
	
	\bibitem[\protect\citeauthoryear{Roy, Banerjee, and Basu}{Roy
		et~al.}{2014}]{roy_tbcas}
	Roy, S., A.~Banerjee, and A.~Basu (2014, Oct.).
	\newblock Liquid state machine with dendritically enhanced readout for
	low-power, neuromorphic vlsi implementations.
	\newblock {\em IEEE Transactions on Biomedical Circuits and Systems\/}~{\em
		8\/}(5), 681--695.
	
	\bibitem[\protect\citeauthoryear{Roy and Basu}{Roy and
		Basu}{2015}]{Roy_unsup_2015}
	Roy, S. and A.~Basu (2015).
	\newblock An online unsupervised structural plasticity algorithm for spiking
	neural networks.
	\newblock {\em IEEE Transactions on Neural Networks and Learning Systems\/}.
	
	\bibitem[\protect\citeauthoryear{Roy, Basu, and Hussain}{Roy
		et~al.}{2013}]{roy_biocas1}
	Roy, S., A.~Basu, and S.~Hussain (2013, Nov).
	\newblock {Hardware efficient, Neuromorphic Dendritically Enhanced Readout for
		Liquid State Machines}.
	\newblock In {\em Proceedings of the IEEE Biomedical Circuits and Systems
		(BioCAS)}, pp.\  302--305.
	
	\bibitem[\protect\citeauthoryear{Roy, San, Hussain, Wei, and Basu}{Roy
		et~al.}{2015}]{Roy_MOT_tnnls_2015}
	Roy, S., P.~P. San, S.~Hussain, L.~W. Wei, and A.~Basu (2015).
	\newblock Learning spike time codes through morphological learning with binary
	synapses.
	\newblock {\em IEEE Transactions on Neural Networks and Learning Systems
		(Revised and resubmitted)\/}.
	
	\bibitem[\protect\citeauthoryear{Schliebs, Fiasch{\'e}, and Kasabov}{Schliebs
		et~al.}{2012}]{Schliebs2012}
	Schliebs, S., M.~Fiasch{\'e}, and N.~Kasabov (2012).
	\newblock Constructing robust liquid state machines to process highly variable
	data streams.
	\newblock In {\em Proceedings of the 22nd International Conference on
		Artificial Neural Networks and Machine Learning (ICANN)}, pp.\  604--611.
	
	\bibitem[\protect\citeauthoryear{Schliebs, Mohemmed, and Kasabov}{Schliebs
		et~al.}{2011}]{Schliebs_ijcnn_2011}
	Schliebs, S., A.~Mohemmed, and N.~Kasabov (2011, Jul.).
	\newblock Are probabilistic spiking neural networks suitable for reservoir
	computing?
	\newblock In {\em Proceedings of the International Joint Conference on Neural
		Networks (IJCNN)}, pp.\  3156--3163.
	
	\bibitem[\protect\citeauthoryear{Sillin, Aguilera, Shieh, Avizienis, Aono,
		Stieg, and Gimzewski}{Sillin et~al.}{2013}]{Sillin2013}
	Sillin, H.~O., R.~Aguilera, H.-H. Shieh, A.~V. Avizienis, M.~Aono, A.~Z. Stieg,
	and J.~K. Gimzewski (2013).
	\newblock A theoretical and experimental study of neuromorphic atomic switch
	networks for reservoir computing.
	\newblock {\em Nanotechnology\/}~{\em 24\/}(38), 384004.
	
	\bibitem[\protect\citeauthoryear{Sporea and Gr\"{u}ning}{Sporea and
		Gr\"{u}ning}{2013}]{Sporea2013}
	Sporea, I. and A.~Gr\"{u}ning (2013, February).
	\newblock Supervised learning in multilayer spiking neural networks.
	\newblock {\em Neural Comput.\/}~{\em 25\/}(2), 473--509.
	
	\bibitem[\protect\citeauthoryear{Thomson, Deuchars, and West}{Thomson
		et~al.}{1993}]{Thomson1993}
	Thomson, A., J.~Deuchars, and D.~West (1993).
	\newblock Single axon excitatory postsynaptic potentials in neocortical
	interneurons exhibit pronounced paired pulse facilitation.
	\newblock {\em Neuroscience 54\/}~(2), 347360.
	
	\bibitem[\protect\citeauthoryear{Varela, Sen, Gibson, Fost, Abbott, and
		Nelson}{Varela et~al.}{1997}]{Varela1997}
	Varela, J., K.~Sen, J.~Gibson, J.~Fost, L.~Abbott, and S.~Nelson (1997).
	\newblock A quantitative description of short-term plasticity at excitatory
	synapses in layer 2/3 of rat primary visual cortex.
	\newblock {\em The Journal of Neuroscience\/}~{\em 17\/}(20), 7926--7940.
	
	\bibitem[\protect\citeauthoryear{Verstraeten, Schrauwen, Stroobandt, and
		Van~Campenhout}{Verstraeten et~al.}{2005}]{Verstraeten}
	Verstraeten, D., B.~Schrauwen, D.~Stroobandt, and J.~Van~Campenhout (2005,
	Sep.).
	\newblock Isolated word recognition with the liquid state machine: a case
	study.
	\newblock {\em Inf. Process. Lett.\/}~{\em 95\/}(6), 521--528.
	
	\bibitem[\protect\citeauthoryear{Wojcik}{Wojcik}{2012}]{Wojcik2012}
	Wojcik, G.~M. (2012).
	\newblock Electrical parameters influence on the dynamics of the
	hodgkinâhuxley liquid state machine.
	\newblock {\em Neurocomputing\/}~{\em 79}, 68--74.
	
	\bibitem[\protect\citeauthoryear{Xue, Hou, and Li}{Xue et~al.}{2013}]{Xue2013}
	Xue, F., Z.~Hou, and X.~Li (2013).
	\newblock Computational capability of liquid state machines with
	spike-timing-dependent plasticity.
	\newblock {\em Neurocomputing\/}~{\em 122\/}(0), 324--329.
	
\end{thebibliography}

\end{document}